\documentclass[10pt,twocolumn,letterpaper]{article}

\usepackage[pagenumbers]{iccv} %

\usepackage[dvipsnames]{xcolor}

\newcommand{\blue}[1]{{\color{blue}#1}}

\newcommand{\linkToPdf}[1]{\href{#1}{\blue{(pdf)}}}
\newcommand{\linkToPpt}[1]{\href{#1}{\blue{(ppt)}}}
\newcommand{\linkToCode}[1]{\href{#1}{\blue{(code)}}}
\newcommand{\linkToWeb}[1]{\href{#1}{\blue{(web)}}}
\newcommand{\linkToVideo}[1]{\href{#1}{\blue{(video)}}}
\newcommand{\linkToMedia}[1]{\href{#1}{\blue{(media)}}}
\newcommand{\award}[1]{\xspace} %

 \usepackage{cuted}
\usepackage{tabularx, colortbl}
\usepackage{multirow}

\definecolor{iccvblue}{rgb}{0.21,0.49,0.74}
\usepackage[pagebackref,breaklinks,colorlinks,citecolor=iccvblue]{hyperref}
\usepackage{bbding}
\usepackage[normalem]{ulem}

\newcommand{\name}{Bayesian Fields\xspace}
\newcommand{\clio}{Clio\xspace}
\newcommand{\lerf}{LeRF\xspace}
\newcommand{\sg}{Semantic Gaussians\xspace}
\newcommand{\ogs}{OpenGaussian\xspace}
\newcommand{\gs}{Gaussian Splat\xspace}
\newcommand{\clip}{CLIP\xspace}
\newcommand{\aib}{Agglomerative Information Bottleneck\xspace}
\newcommand{\fsam}{FastSAM\xspace}
\newcommand{\coco}{MSCOCO-Captions\xspace}
\newcommand{\cg}{ConceptGraphs\xspace}
\newcommand{\thres}{task\xspace}
\newcommand{\batch}{batch\xspace}
\newcommand{\myParagraph}[1]{{\bf #1.}}
\newcommand{\real}{\mathbb{R}}
\renewcommand{\ie}{i.e.,\xspace}
\renewcommand{\eg}{e.g.,\xspace}

\def\paperID{6340} %
\def\confName{ICCV}
\def\confYear{2025}

\title{\name: Task-driven Open-Set Semantic Gaussian Splatting}

\author{Dominic Maggio \quad Luca Carlone\\
Laboratory for Information \& Decision Systems, Massachusetts Institute of Technology\\
{\tt\small \{drmaggio, lcarlone\}@mit.edu}
}

\begin{document}
\twocolumn[
{
\renewcommand\twocolumn[1][]{#1}%
\maketitle
\begin{center}
\centering
\includegraphics[width=\textwidth]{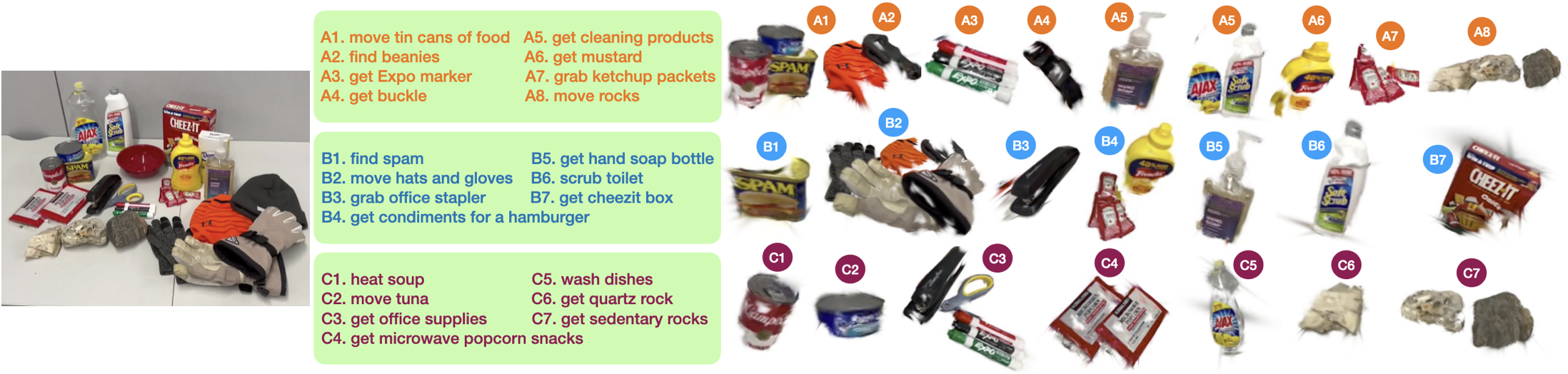}\vspace{-2mm}
\captionof{figure}{\name takes a collection of posed RGB images and a task list and extracts objects at the right granularity. 
Given a set of tasks in natural language 
such as ``move hats and gloves'' (B2), \name considers a pile of clothing collectively as one object in its map. 
Providing more specific tasks such as ``find beanies'' 
and ``get buckle'' (A2, A4) results in a more fine-grained scene representation at 
the correct granularity to support the tasks. We also detect different instances of objects relevant for the same task. For example, 
there are two separate clusters of cleaning products corresponding to ``get cleaning products'' (A5), since these objects are 
spatially separated in the scene. 
Objects shown are the 3D extracted objects that \name estimates are the most relevant for each task. 
}
\label{fig:cover_fig}
\end{center}
}]

\begin{abstract}
Open-set semantic mapping requires (i)
determining  
the correct granularity to represent the scene (\eg how should objects be defined), and (ii) 
fusing semantic knowledge across multiple 2D observations into an overall 3D reconstruction  
---ideally with a high-fidelity yet low-memory footprint. 
While most related works bypass the first issue by %
grouping together primitives with similar semantics (according to some manually tuned threshold), 
 we recognize that the object granularity is task-dependent, and develop a task-driven semantic mapping approach.
To address the second issue, current 
practice is to average visual embedding vectors over multiple views. Instead, we show the benefits of using
 a probabilistic approach based on the properties of the underlying visual-language foundation model, and leveraging Bayesian updating 
to aggregate multiple observations of the scene. 
The result is
 \emph{\name}, a task-driven and probabilistic approach for open-set semantic mapping.
 To enable high-fidelity objects and a dense scene representation, 
\name uses 3D Gaussians which we cluster into task-relevant objects, allowing for both easy 3D object 
extraction and reduced memory usage. We release \name open-source at \url{https://github.com/MIT-SPARK/Bayesian-Fields}.
\end{abstract} %
\vspace*{-38pt}
\section{Introduction}
\label{sec:intro}

A fundamental problem in creating an open-set semantic environment representation is determining what is the right granularity 
to represent a scene and how should objects be defined. 
This problem was formalized by Marr in his seminal work~\cite{Marr83book}
and most researchers now would agree that the correct scene representation depends on the tasks an agent 
must complete~\cite{Soatto16iclr-visualRepresentations}. 
For example, if winter gear placed on top a table should be considered a 
collective object of clothing or as multiple objects for each clothing type, or as buttons, buckles, and fabric, 
depends on whether a user needs to 
simply move the pile or needs to repair a buckle on an article of clothing; see~\cref{fig:cover_fig} for other examples. 

Classical computer vision tasks, such as determining the geometry or color of a point in a scene, 
can be done by considering an infinitely small surface around the point and either computing the surface 
normal or estimating light 
emittance. However, reasoning over semantics requires going beyond points.
Indeed, in the limit of considering an infinitesimal point, semantics becomes unobservable as objects are split into atoms, electrons, perhaps stopping only at  
quarks or free space of nothingness. 
Instead, to ask what the semantics of a particular point is, one must first arbitrarily define a region 
of interest. For example, pointing at the top of a mustard bottle and asking what is the semantics 
would likely either imply wanting the semantics to be of the cap or of the entire bottle. 
Much beyond being an interesting philosophical discussion, these questions are critical in order to 
design an effective semantic map. A map that is too fine-grained might miss high-level 
concepts (in addition to having impractically high memory requirements), 
while a map that is too coarse might ignore important semantic concepts. Since a map's usefulness is measured by 
its ability to support the tasks of its user, we investigate task-driven mapping 
where the granularity of the map is chosen to 
support a set of tasks specified in natural language (\cref{fig:cover_fig}).

Recently, \clio~\cite{Maggio24ral-clio} has shown how to leverage advances in 
open-set vision-language foundation models (e.g., CLIP~\cite{Radford21icml-clip}) to execute task-driven 
mapping in practice, given natural language tasks. 
Towards this goal, \clio first over-segments a scene into 3D object primitives represented as meshes, and then leverages 
tools from information theory~\cite{Slonim99nips-AgglomerativeIB} to cluster the primitives into task-relevant objects. 

One challenge with Clio~\cite{Maggio24ral-clio}  and related approaches~\cite{Gu24icra-conceptgraphs, Qin24cvpr-langsplat, Guo24arxiv-semanticGS} is that the relevance of an object is measured directly using the cosine 
similarity between visual and text embeddings, and the results often rely on 
tuned thresholds to isolate relevant objects. 
However, practitioners are well-aware that cosine similarities from models such as CLIP typically produce values in a small range, which makes setting thresholds challenging in practice. 
Moreover, it has been shown that visual and text embeddings from CLIP form disjoint sets~\cite{Hitchcox22ral-robustEstimation, Mistretta25icml-crossTheGap}, making direct comparisons harder.

A second open challenge in the current literature is how to properly fuse semantic observations from 2D foundation models across multiple views. 
The majority of current works simply average semantic 
embeddings~\cite{Maggio24ral-clio, Gu24icra-conceptgraphs, Guo24arxiv-semanticGS, Qin24cvpr-langsplat, Wu24neurips-opengaussian}, 
with some also attempting to approximate the uncertainty of semantic measurements~\cite{Shi24cvpr-LEGaussians, Wilson24arxiv-gsUncertainty}. 
Ideally, we want a more grounded way to probabilistically fuse observations over time that accounts for uncertainty and differences across observations of 
a scene, such as that object semantics can appear different from different viewpoints~\cite{Poggio21book-machineObjects}.

A third challenge is that representing 3D object primitives with 3D noisy meshes as in~\cite{Maggio24ral-clio} limits the ability of the approach to 
produce accurate geometry and 
high-fidelity photometric rendering. Gaussian Splatting~\cite{Kerbl23Ttog-GaussianSplatting} has quickly 
become a popular alternative scene representation 
but current work that incorporates semantics into Gaussian splatting is met with problems of excessive 
memory, since attaching high dimensional feature 
embeddings is only feasible for small scenes~\cite{Guo24arxiv-semanticGS}. Thus, works typically 
either compress the features using a small scene-specific network ---which requires additional training and makes scene 
updating difficult~\cite{Qin24cvpr-langsplat}--- or quantizes all features of a scene into a 
set of representative features~\cite{Shi24cvpr-LEGaussians} ---which does 
not consider the task-driven granularity of the scene.  

\myParagraph{Contributions} We provide the following key contributions to address the three challenges discussed above:
\begin{itemize}
    \item Our first contribution (\cref{sec:clip_prob}) is a probabilistic interpretation of vision-text similarity. In particular, we  
    provide a grounded way to relate cosine similarities between visual observations and text-based task descriptions 
    with the probability that an object is relevant for a task.
    \item Our second contribution (\cref{sec:measuresment model}) is a measurement model of 
    semantics that accounts for uncertainty and differences in multiple semantics observations of 
    a scene. This ---together with our probabilistic interpretation of vision-text similarity--- allows us to combine 
    semantic knowledge across views using Bayesian updating. 
    \item Our third contribution (\cref{sec:get_objects}) is a method to construct a task-driven, memory efficient, semantic Gaussian Splatting representation by 
    assigning the 3D Gaussians probabilities of being relevant to a set of tasks and clustering the Gaussians into task-relevant objects using 
    the Information Bottleneck~\cite{Tishby01accc-IB} principle as done in~\cite{Maggio24ral-clio}. 
\end{itemize}

We call the resulting approach \emph{\name} (\cref{fig:cover_fig}).
Given a Gaussian Splat, \name segments the scene into task-relevant objects,
runs in minutes (~\cite{Wu24neurips-opengaussian} takes over an hour), and does not require further training.
Our work joins recent efforts in allowing for 3D object extraction from the Gaussian Splat~\cite{Wu24neurips-opengaussian, Shi24cvpr-LEGaussians}, instead 
    of focusing on rendering 2D semantics maps~\cite{Qin24cvpr-langsplat, Yu24arxiv-legs}.

\section{Related Work}
\label{sec:related_work}

\myParagraph{Visual Foundation Models} Work in class-agnostic image segmentation~\cite{Kirillov23iccv-SegmentAnything,Zhao23Arxiv-FastSam, zhang23arxiv-mobile_sam} and 
vision-language 
models~\cite{You23arxiv-ferret, Zhang24arxiv-ferretv2, Radford21icml-clip, Oquab23arxiv-dinov2, Chen24arxiv-Cloc, 
Vasu24arxiv-fastvlm} has 
led to rapid advances in semantic understanding of 2D images~\cite{Liu23arxiv-groundingDino, Wang23arxiv-samClip}. 
Recent work uses vision-language models for 2D object detection~\cite{Li22iclr-lseg, Minderer22eccv-owlVit} and 
3D scene understanding~\cite{Gu24icra-conceptgraphs, Jatavallabhula23rss-ConceptFusion, Huang24neurips-chatscene, Wang23arxiv-chat3D, Shafiullah22arxiv-clipFields, 
Ye24eccv-gaussianGrouping, Lyu24arxiv-gaga}. 
Related works propose open-set SLAM systems~\cite{Zhu24cvpr-sniSlam, Pham24arxiv-GoSlam, Li24arxiv-HiSlam} and 
use open-set landmarks for localization~\cite{Peterson24arxiv-Roman}.
A study of practices of 
implementing CLIP for 3D scene understanding is given in~\cite{Kassab24arxiv-BareNecessities}.
The theory of the structure of the embedding space of visual-text embedding 
is studied in~\cite{Hitchcox22ral-robustEstimation, Park23arxiv-llmGeometry, Qi24arxiv-latentSpaceControl, Park24arxiv-llmHierachicalGeometry}, 
and limitations of CLIP~\cite{Radford21icml-clip} are discussed in~\cite{Lewis22arxiv_clipLimitation, Tong24cvpr-eyesWideShut}.

\myParagraph{Open-Set Semantics with Radiance Fields} In the past few years, NeRF~\cite{Mildenhall21acm-nerf} and Gaussian Splatting~\cite{Kerbl23Ttog-GaussianSplatting} have 
become popular scene representations for applications ranging from SLAM~\cite{Matsuki24cpvr-gsSlam} to 
pose estimation monitoring~\cite{Maggio23ral-VERF}. 
Multiple works have extended NeRF and Gaussian Splatting to be able to render semantic embeddings~\cite{Kerr23iccv-lerf, 
Qin24cvpr-langsplat, Yu24arxiv-legs, Zuo24ijcv-fmgs}. Like ours, some recent works instead also allow for 3D object 
extraction~\cite{Wu24neurips-opengaussian, Shi24cvpr-LEGaussians, Guo24arxiv-semanticGS}. 
\sg~\cite{ Guo24arxiv-semanticGS} assigns CLIP vectors to each Gaussian and averages the embedding vectors across multiple observations. 
This leads to high memory usage and is prone to noisy object extraction as Gaussians are not grouped into 3D objects. \ogs~\cite{Wu24neurips-opengaussian} 
trains instance features and a codebook given multi-resolution 2D masks, to cluster 3D Gaussians which produces cleaner extraction of 3D objects. 

\myParagraph{Determining the Granularity of a Scene Representation} The mentioned open-set mapping 
work empirically defines the granularity of the objects in the scene by manually setting thresholds of semantic similarity.
The topic of how to optimally represent objects and define a scene has also been the subject of theoretical discussion in 
multiple works~\cite{Soatto16iclr-visualRepresentations, Parameshwara23arxiv-visualFoundationModels, Marr83book, Poggio21book-machineObjects, 
Simon96book-scienceOfArtificial, Li17cvpr-objectSaliency}. 
Most similar to the proposed \name is \clio~\cite{{Maggio24ral-clio}}. 
\clio uses the Information Bottleneck principle~\cite{Tishby01accc-IB, Slonim99nips-AgglomerativeIB} to form an open-set task-relevant metric-semantic map. 
As mentioned in \cref{sec:intro}, our improvements over \clio are threefold: a more grounded approach 
to determine object relevance from observation of semantic embeddings, a measurement model that allows for 
Bayesian fusion of semantic embeddings across multiple views, 
and a task-driven framework that uses Gaussian Splatting in place of meshes for representing objects. 

Similar to our question of selecting the correct granularity of a scene, Garfield~\cite{Kim24cvpr-garfield} trains a scale-conditioned 
affinity field for a NeRF scene, based on multiple scales of SAM~\cite{Kirillov23iccv-SegmentAnything} masks. The approach can then 
be used with Gaussian Splatting by querying the affinity field at the Gaussian centers.
However, it sets granularity using a user-defined knob, which indiscriminately changes the granularity of the entire scene.
Here, we want to automatically pick the correct scene granularity and importantly recognize that some parts 
of the scene may require a coarse representation 
while others need a finer representation. While Garfield does not include semantics in its map, we provide qualitative comparisons 
with \name in \cref{sec:garfield}.

\section{Problem Formulation and Preliminaries}
\label{sec:notation}

\begin{figure*}[ht]
  \centering
  \includegraphics[width=\linewidth]{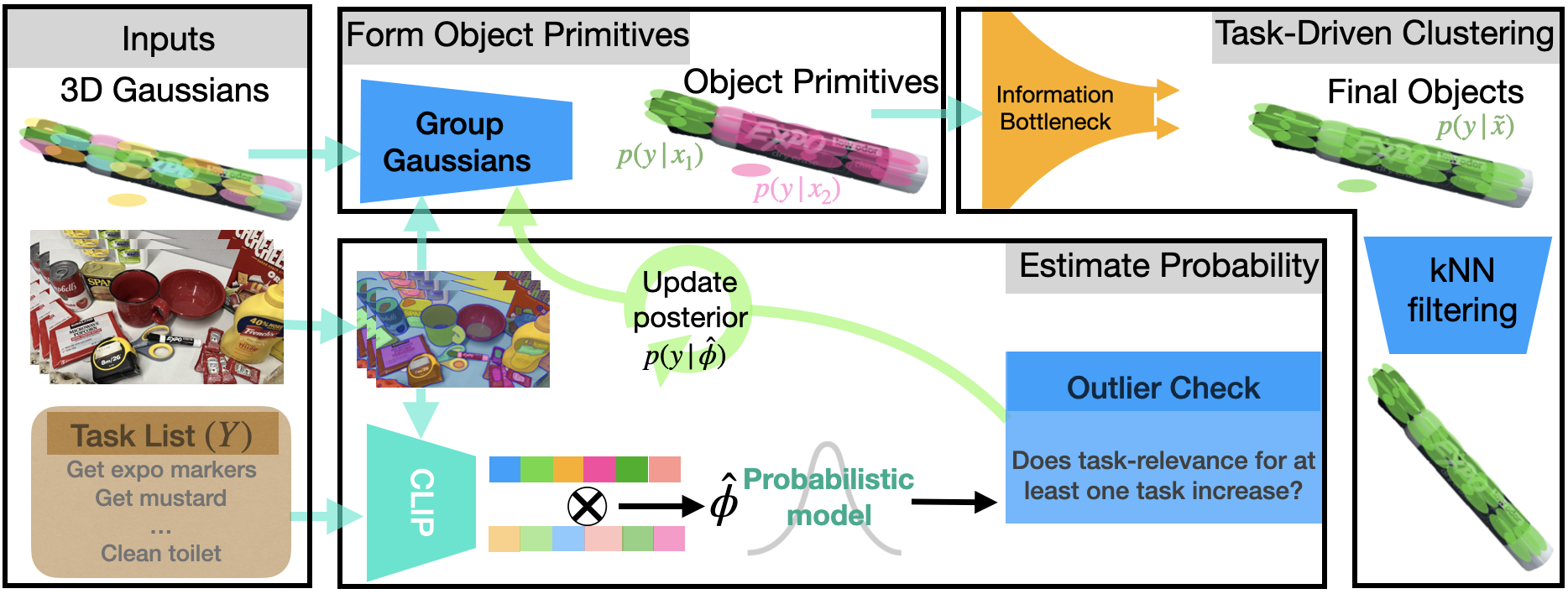}
  \caption{Overview of \name. Our system takes in a trained 3D Gaussian Splat, RGB images with poses, and a list of tasks. We compute cosine scores 
  from CLIP embeddings of masked image segments and the task list and convert them to probabilities which are aggregated across
  multiple observations using our measurement model and Bayesian updating. Gaussians are grouped into 3D primitives and then 
  clustered into task-relevant objects. Point Cloud filtering (kNN) is applied as a final step to remove potential floaters.}
  \label{fig:pipeline}
\end{figure*}

\myParagraph{Problem Statement}
We adopt the problem statement of \clio~\cite{Maggio24ral-clio} where we are given a set of tasks $Y$ in natural language from the user (\cref{fig:cover_fig}), and
the objective is to form a map representation that contains a set of objects $\tilde{X}$ at the correct 
granularity to support the tasks, and then allows for extracting the most relevant objects for each task.
In this paper, we additionally assume a 3D Gaussian Splat of the scene is available along with RGB images and their poses.

\myParagraph{Task-Driven Objects via the Information Bottleneck}
\clio shows that the problem statement can be formalized mathematically using the classical Information Bottleneck theory~\cite{Tishby01accc-IB}. 
\clio first over-segments the scene into a set of 3D primitives $X$, and then clusters the primitives 
into task-relevant objects $\tilde{X}$, while preserving information about the set of tasks $Y$. 
Writing soft assignments in terms of probabilities $p(\tilde{x}|x)$ (\ie the probability that a primitive $x \in X$ is clustered into an object $\tilde{x} \in \tilde{X}$), the task-driven clustering problem can be written using the Information Bottleneck as: 
\begin{equation}
	\label{eq:ib}
	\textstyle\min_{p(\tilde{x}|x)} I(X;\tilde{X}) - \beta I(\tilde{X}; Y),
\end{equation}
where $\beta$ controls the trade-off between compression and preserving information about the tasks. 
The Information Bottleneck requires defining $p(y|x)$, which in our problem is the relevance between a primitive $x$ 
and one of the given tasks $y$. 
The optimization problem~\eqref{eq:ib} can be solved incrementally using the \aib algorithm~\cite{Slonim99nips-AgglomerativeIB},
 which defines a graph where nodes correspond to primitives, and edges connect nearby primitives, and then
 gradually clusters primitives together.
This produces a hard clustering assignment $p(\tilde{x}|x)$. More in detail, each edge $(i,j)$ is assigned a weight which measures the 
amount of distortion in information caused by merging primitive clusters $\tilde{x}_i$ and $\tilde{x}_j$ as:
\begin{equation}
	\label{eq:aib_weight}
	d_{ij} = (p(\tilde{x}_i) + p(\tilde{x}_j)) \cdot D_{\textrm{JS}}[p(y|\tilde{x}_i), p(y|\tilde{x}_j)],
\end{equation}
where $p(\tilde{x}_i)$ and $p(\tilde{x}_j)$ are computed by the algorithm
(and intuitively store the number of primitives merged in  
clusters $i$ and $j$), and $D_{\textrm{JS}}$ is the Jensen-Shannon divergence. At each iteration, edges are merged greedily based on their weight.

\clio~\cite{Maggio24ral-clio} uses the cosine score (which is not a probability) as a proxy for $p(y|x)$ and temporally 
associates semantic observations of primitives (represented as meshes) by averaging semantic embedding vectors across views. 
Below, we design a probabilistically grounded approach to determine task-relevance probabilities from observations of semantic 
embeddings and to aggregate measurements over multiple views instead of taking simple averages.

\section{\name: A Probabilistic Approach to Task-Driven Semantic Mapping}
\label{sec:method}

Here we present the design of \name. Given a trained \gs, list of tasks $Y$ in 
natural language, and RGB images with poses, \name produces a 3D map of task-relevant objects represented with Gaussian Splatting (\cref{fig:pipeline}). 
Using a visual-language foundation model (\clip~\cite{Radford21icml-clip}), we compute embedding vectors for each task. For 
each new RGB image, \name uses a class agnostic segmentation network (\fsam~\cite{Zhao23Arxiv-FastSam}) to get a set of masked 2D regions, {$L$}, 
and computes \clip vectors for each masked region. 
We generate fine masks to intentionally over-segment the scene, and masks are filtered such that each pixel per image is assigned to at 
most one mask. 
\name then uses a probabilistic model 
to extract the probability of the observation being task-relevant for each of the tasks from the corresponding semantic embeddings (\cref{sec:clip_prob}), and aggregates information across frames using Bayesian updating (\cref{sec:measuresment model}). 
Concurrently, the 2D segments are 
mapped to 3D Gaussians, which store the aggregated probabilities, and the 3D Gaussians are assigned to over-segmented
3D object primitives. The over-segmented 3D primitives are finally clustered into 
task-relevant objects using the \aib (\cref{sec:get_objects}); $k$-nearest neighbor filtering is applied over all Gaussians in each 
object to remove potential floaters.

\subsection{Getting Probabilities of Relevance from CLIP}
\label{sec:clip_prob}
Given a set of tasks, $Y$, we desire to determine the probability that a visual embedding vector $f_\text{img} \in \real^D$ 
is associated with a text embedding $f_{\text{task}} \in \real^D$ of a particular task $y \in Y$. For simplicity, assume throughout that the 
embeddings are always normalized.
Common practice of comparison is to just compute the cosine score, $\phi(f_\text{img}, f_{\text{task}}) = f_\text{img}^\top f_{\text{task}}$. 
However, the cosine score typically falls in a narrow range of values that contain information about 
relative comparison (i.e., larger values imply more similarity between embedding vectors than do smaller values), 
but do not directly provide an informative measure of the likelihood that embedding vectors are associated.

In this section, given a cosine score, $\phi$, we want to determine $p(y=1 | \phi)$, where each task, $y$, has value 1 if it is 
relevant for the task and 0 otherwise. This binary decision is similar to how CLIP is trained via a 
contrastive loss, where image and text are labeled as either being associated or disassociated and embeddings are 
trained to maximize $\phi$ in the case of the former and minimize for the latter.

\begin{figure}[h]
  \centering
  \begin{subfigure}[t]{0.48\linewidth}
    \includegraphics[width=\linewidth]{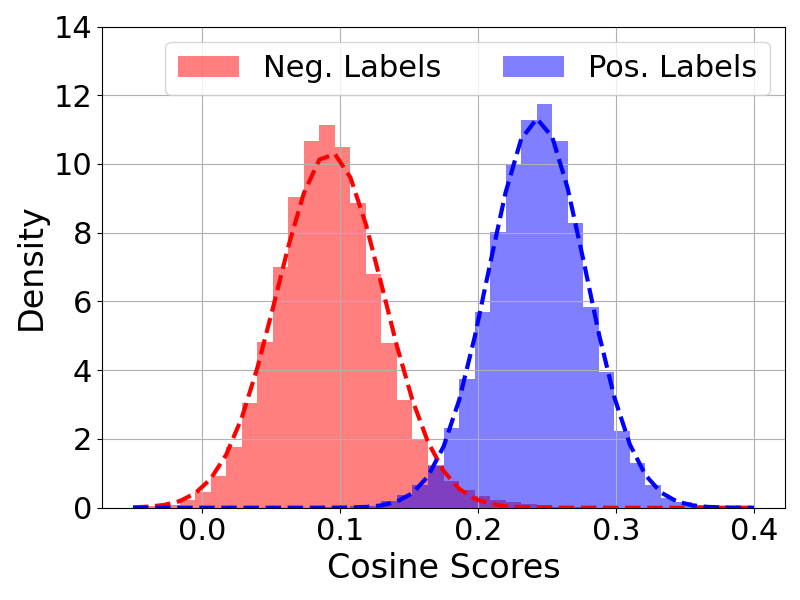}
    \caption{Normalized distributions of the cosine scores of 32,000 positive labels and 992,000 negative labels 
    from the \coco dataset. A best fitting Gaussian pdf is plotted for each, which for the negative and positive distributions 
    has a mean and standard deviation of 0.092, 0.039 and 0.243, 0.035 respectively. }
    \label{fig:gaussian_yes_no}
  \end{subfigure}
  \hfill
  \begin{subfigure}[t]{0.48\linewidth}
    \includegraphics[width=\linewidth]{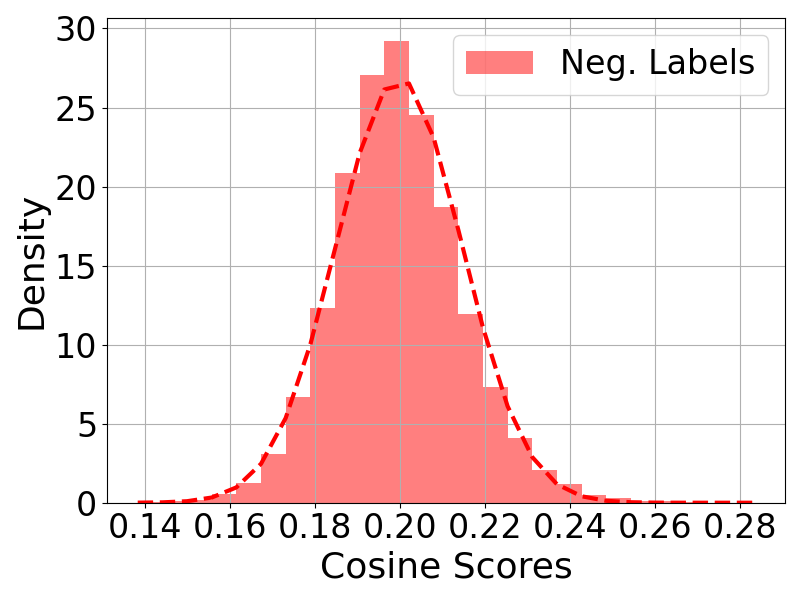}
    \caption{Normalized distributions of the cosine scores of about 28,000 negative labels from 
    the \clio~\cite{Maggio24ral-clio} Apartment scene, obtained by generating image 
    embeddings of masked image segments from \fsam using \name processing. 
    A best fitting Gaussian pdf is plotted which has a 
    mean and standard deviation of 0.20, 0.015.}
    \label{fig:gaussian_np}
  \end{subfigure}
  \caption{Gaussian behavior of cosine scores for CLIP image and text embeddings.}
  \label{fig:single-column-figure}
\end{figure}

Using a dataset of images and labels, in our case \coco~\cite{Lin14eccv-coco} which has text descriptions 
for each image such as ``A large bird sitting up in a tree branch next to a statue'', we compute {pairwise cosine 
scores between batches of image and text embeddings}. We collect all cosine scores for the positive and negative 
labels and visualize the normalized distribution of each in \cref{fig:gaussian_yes_no}. 
\emph{A surprising observation that arises from the  figure is that 
these two distributions can be fitted by a nearly perfect Gaussian distribution with similar standard deviation.} 
These are the distributions for $p(\phi|y=1)$ and $p(\phi|y=0)$. We suspect this behavior is caused by a narrow cone of embedding space for image and text embeddings known as the cone effect~\cite{Hitchcox22ral-robustEstimation}, or is induced by the loss used for training. 

Since the cosine score can be modeled effectively with a Gaussian distribution, we can now compute $p(y=1 | \phi)$ (the probability that an observation is relevant for task $y$) using Bayes' Rule as follows: 
\begin{equation}
  p(y=1 | \phi) = \frac{p(\phi|y=1) p(y=1)}{p(\phi|y=0) p(y=0) + p(\phi|y=1) p(y=1)},
  \label{eq:bayes}
\end{equation}
where $p(\phi|y)$ can be estimated from the empirical distributions (as the one in \cref{fig:gaussian_yes_no}), and 
$p(y=0)$ and $p(y=1)$ can be set to be equal (\ie uninformative priors since no prior about the probability of relevance is available). %

\hspace{-5mm}

\subsection{Measurement Model for CLIP and Fusing Probabilities Across Multiple Observations}
\label{sec:measuresment model}

\begin{figure}[h]
  \centering
  \begin{subfigure}[t]{0.35\linewidth}
  \hspace{-2mm}
    \includegraphics[width=\linewidth]{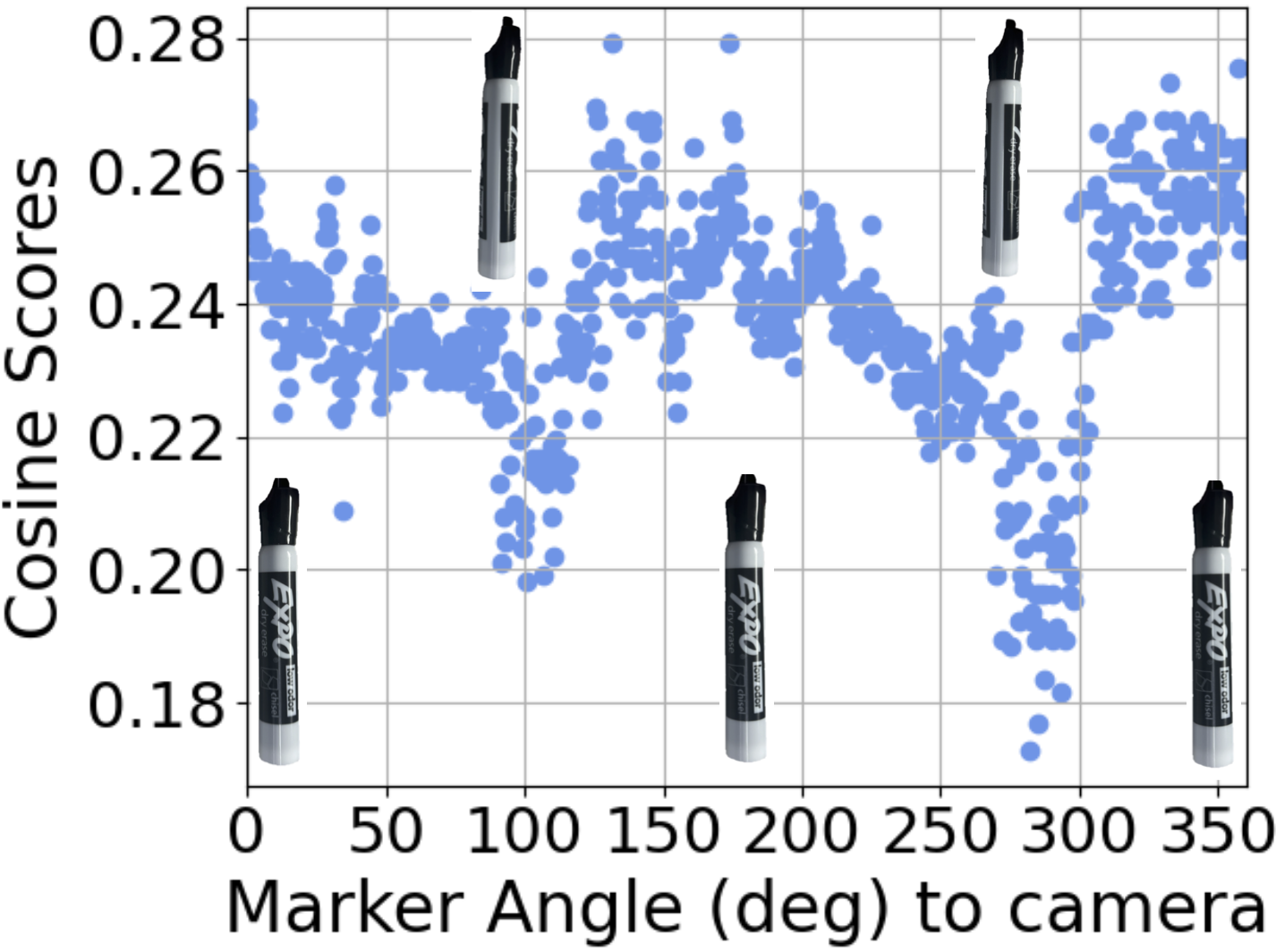}
    \caption{Cosine score to ``get Expo marker'' as the camera is fixed and the marker spins on a turntable. Visuals show 
    views of the marker at different angles, where cosine score is highest when the text \textit{Expo} is visible.}
    \label{fig:sinusoid}
  \end{subfigure}
  \hfill
  \begin{subfigure}[t]{0.33\linewidth}
    \includegraphics[width=\linewidth]{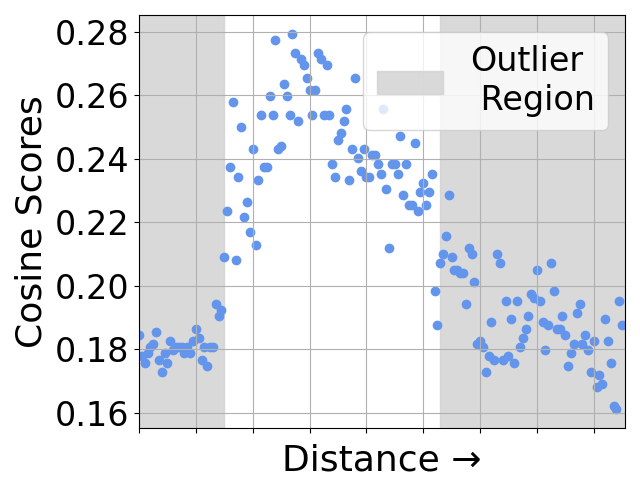}
    \caption{Cosine score to ``get tape measure'' as the tape measure goes from being too close to too far from the camera. 
    Distance regions producing outlier cosine scores are approximately highlighted in grey.}
    \label{fig:outlierPlot}
  \end{subfigure}
  \hfill
  \begin{subfigure}[t]{0.27\linewidth}
    \includegraphics[width=\linewidth]{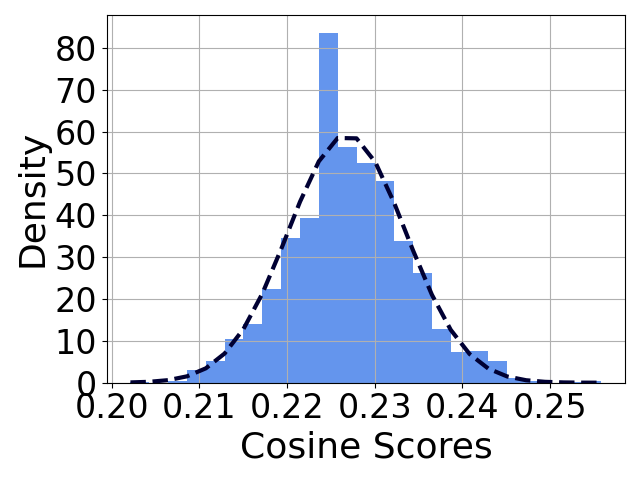}
    \caption{Gaussian noise in cosine score to ``get tape measure'' for fixed camera and tape measure.
    Best fitting Gaussian distribution is plotted.}
    \label{fig:gaussian_noise}
  \end{subfigure}
  
  \caption{Demonstration of three types of error in cosine score between text and multiple masked 
  frames of the object --- an Expo marker for (a) 
  and a tape measure for (b) and (c). 
  The object is masked with SAM2~\cite{Ravi24arxiv-sam2} for accurate tracking across frames.}
  \label{fig:cosine_errors}
\end{figure}

\Cref{eq:bayes} provides a way to compute $p(y | \phi)$ given a single observation with cosine score of $\phi$.
However, in practice (i) we make multiple observations of the same region, and (ii) these observations 
are likely to produce different values of cosine scores.
In \cref{fig:cosine_errors}. we describe 
three reasons why the observed value $\hat{\phi}$ will differ from an ``oracle'' $\phi$, where $\phi$ should be the same for all multi-view observations 
of the same object. 
In particular, the figure provides empirical evidence that the cosine score depends on the 
angle of the object with respect to the observer, the distance to the observer, and white noise from the foundation model.

To address the challenge that the observation of cosine score is noisy and view-dependent,  
we first write a model relating the observed cosine score $\hat{\phi}$ to the ``oracle'' score $\phi$:
\begin{equation}
  \hat{\phi} = \phi + f(\theta, z) + \epsilon;\quad \epsilon  \sim \mathcal{N}(0, \sigma^2),
  \label{eq:measurement_model}
\end{equation}
where %
$f(\theta, z)$ is a function that 
depends on the angle $\theta$ between the object and the camera
 (\cref{fig:sinusoid}) and their distance $z$ (\cref{fig:outlierPlot}), 
and $\epsilon$ is zero-mean Gaussian noise (\cref{fig:gaussian_noise}).

Now, we can replace \cref{eq:bayes} with the following Bayesian update rule, which computes probability of 
relevance by fusing information across multiple frames, while accounting for the fact that we measure $\hat{\phi}$ instead of $\phi$:
\begin{equation}
  p(y=1 | \hat{\phi}_{1:t}) = \frac{p(\hat{\phi}_t | y=1) p(y=1 | \hat{\phi}_{1:t-1})}{p(\hat{\phi}_{t})}.
  \label{eq:bayes_filter}
\end{equation}
\Cref{eq:bayes_filter} computes the task relevance for all the observations of an object collected between time $1$ and the current time $t$, 
given the probability at time $t-1$, namely $p(y=1 | \hat{\phi}_{1:t-1})$, and the newly acquired observation $\hat{\phi}_t$. In other words,~\cref{eq:bayes_filter} allows us to recursively update the probability of relevance
of an object given newly acquired observations (\cref{fig:pipeline}, bottom).
In \cref{eq:bayes_filter}, $p(\hat{\phi}_t | y=1)$ is computed from the empirical distributions (\cref{fig:single-column-figure})
 but with some extra care for measurement errors (as described below), while 
 $p(\hat{\phi}_{t})$ is 
the marginal probability computed as $p(\hat{\phi}_t | y=1) p(y=1 | \hat{\phi}_{1:t-1}) + 
p(\hat{\phi}_t | y=0) p(y=0 | \hat{\phi}_{1:t-1})$, where we assume the noise at time $t$ to be independent from the 
noise at the previous times; the term $p(y=1 | \hat{\phi}_{1:t-1})$ is recursively updated by~\eqref{eq:bayes_filter} and initialized to   
0.05 (\ie a low probability of being task-relevant). 

{\bf Using~\eqref{eq:measurement_model} to Account for Measurement Errors.}
{In order to get $p(\hat{\phi}|y)$ using the empirical distribution $p(\phi|y)$, we make the following amends 
based on the measurement model in \eqref{eq:measurement_model}. 
Firstly, our use of \fsam will tend to form multiple primitives (as described in \cref{sec:get_objects}) for 
different sides of a potential object, making the influence of the angle $\theta$ on each primitive small in practice.
Furthermore, if $z$ is in a range of observable values, we can assume Gaussian noise captured by $\epsilon$. 
However, if $z$ is a distance either too far or too close (such as revealing an incomprehensible occluded view of an object), 
the value is an outlier, as demonstrated in \cref{fig:outlierPlot}. 
To mitigate potential negative effect from outliers,}
we develop a strategy ablated 
to be effective in \cref{sec:ablations} as follows:
if the posterior probability in~\cref{eq:bayes_filter} decreases for all tasks, we do not update the probability, otherwise, 
we apply a Bayesian update. Intuitively, measurements that are 
more likely (given by $p(\hat{\phi} | y=1) < p(\hat{\phi} | y=0)\; \forall y \in Y$)
to be irrelevant objects do not need to be updated and can maintain a low probability. 
High probability values get updated, and primitives that are truly relevant should be measured with multiple high probabilities, 
while outliers giving false low probabilities are discarded. With this procedure, $p(\hat{\phi} | y)$ needed for \cref{eq:bayes_filter} is 
equivalent to $p(\phi | y)$ with added variance from the distribution of $\epsilon$. 

\subsection{Getting Task-Relevant Objects}
\label{sec:get_objects}
\myParagraph{Forming Object Primitives}
We first group 3D Gaussians into (over-segmented) 3D primitives $X$ (\cref{fig:pipeline}, top center); these will be then fed to the Information Bottleneck to extract task-relevant objects. %
More in detail, for each set of masked 2D regions $L$ (obtained using \fsam), similar to \sg~\cite{Guo24arxiv-semanticGS}, we compute the Gaussians that are in 
view of the mask and determine which ones are not occluded by comparing their 3D centroids to depth images rendered 
from the Gaussian Splat, resulting in a set of Gaussians {$G_{L_i}$ associated to each masked 2D region $L_i$.} %

For each set of Gaussians $G_{L_i} \in G_L$, any of {the Gaussians in the set that have not previously been assigned 
to a 3D primitive} 
are assigned to a new primitive. %
For all the Gaussians in $G_{L_i}$ that {have a previous assignment}, they keep their 
same assignment if {the number of Gaussians in their assigned primitive is less than the number 
of Gaussians in $G_{L_i}$}, otherwise they are 
assigned to the new primitive. This allows us to {over-segment} the scene {over time} into a set of fine-grained 3D primitives. 

We compute the CLIP vector $f_\text{img}$ of each image segment in $L$ and 
update the posterior probability of each Gaussian in $G_{L_i}$ using \cref{eq:bayes_filter}.
As an added speedup, if all Gaussians in $G_{L_i}$ are outliers rejected using the protocol in \cref{sec:measuresment model},  
we do not run the primitive assignment routine on $G_{L_i}$ which reduces the overall number of 3D primitives.

Finally, for all 3D primitives, $x \in X$, we compute $p(y|x)$ by averaging the probability vectors $p(y|\hat{\phi})$ 
from each Gaussian in that primitive (a toy example of a marker with two object primitives and corresponding $p(y|x)$ 
is shown in \cref{fig:pipeline}). This is reasonable since being assigned to the same primitives implies having the 
same specified semantic granularity and hence the same $p(y|\phi)$. We then add a null task which is the probability 
that a primitive is relevant for no tasks. %
Lastly, we normalize $p(y|x)$. 

\myParagraph{Creating Primitives Edges} Given that we now have a set $X$ of 3D primitives and $p(y|x)$, the final step 
before running the \aib is to establish edges connecting primitives. Only connected primitives are eligible to be merged. 
We add edges between primitives if the smallest 3D bounding boxes 
containing the centroids of all Gaussians for a pair of primitives has non-zero overlap. 
Intuitively, this only allows merging primitives that are near each other.
Note that since some connected primitives may have identical task 
probabilities (and hence zero Jensen-Shannon divergence \cref{eq:aib_weight}) due to maintaining their initial prior 
probabilities, we can automatically merge these to reduce computation. 

\myParagraph{Selecting Most Relevant Objects}
 Given the graph of 3D primitives, we can run the \aib algorithm (\cref{fig:pipeline}, top right).
The \aib effectively clusters the 3D primitives into 
tasks-defined objects $\tilde{X}$, and, for each object, it computes $p(y|\tilde{x})$, which is the probability of that object being relevant for each given task. 
To further compress the scene, we can optionally remove objects from $\tilde{X}$ that are unlikely to 
be task-relevant (these objects are typically large clusters of task-irrelevant primitives) by discarding objects that do not have a 
probability greater than 0.1 to any of the tasks. 
For a particular task, $y$, the most relevant $k$ objects are those with the highest probabilities for the task, 
given as: 
\begin{equation}
  \{ \tilde{x}_1, \tilde{x}_2, \dots, \tilde{x}_k \} = \arg\max_{\tilde{x}} \, p(y=1 | \tilde{x}).
  \label{eq:select_object}
\end{equation}
An alternative is to compute a CLIP vector for each object by rendering a view of the object 
(chosen as the training pose with highest visibility of the object) from the 3D Gaussian Splat and 
selecting objects based on highest cosine score to a text embedding of the task. 
However, we find this more complicated alternative does not improve performance 
as demonstrated in \cref{sec:ablations}. %

\section{Implementation Details}
\label{sec:implementation}

We build our system on top of Semantic Gaussians~\cite{Guo24arxiv-semanticGS} and run \name 
with an RTX 3090 GPU and Intel i9-12900K CPU. Each 3D Gaussian Splat is trained for 30,000 iterations. 
Since floaters in the Gaussian Splat and noisy masks can result in 3D primitives that contain outlier Gaussians, 
we perform $k$-nearest neighbor filtering on the Gaussian centroids of the task-relevant objects
to remove noisy Gaussians. We ablate this choice in \cref{sec:ablations}. As described in \cref{sec:get_objects}, 
for all experiments we remove objects from the set of clustered objects that have a probability of relevance below 0.1 for 
every task in the list.
We run \name on every frame to demonstrate low mapping time without need to downsample the frames.

Since the \coco images are full images and the text captions are rich descriptions, while we use 
masked image segments often at low resolution, we calibrate the Gaussian behavior of cosine scores on a 
dataset with our configuration. We can easily find $p(\phi|y=0)$ by using a set of task labels 
that have no relevant objects in the scene. We visualize the results in \cref{fig:gaussian_np} and again confirm 
our observation that the distribution can be modeled by a well-fitted Gaussian. In this case, the mean is 
substantially greater than that of \coco (0.09 vs 0.20) which is expected due to the large difference in configuration. 
We thus allow ourselves to select a mean for the 
Gaussian distributions, for which we use the 
same standard deviation of 0.035 determined from \coco along with 0.20 for $p(\phi|y=0)$ and experimentally determine 0.27 
to perform the best for $p(\phi|y=1)$.  %

\section{Experiments}
\label{sec:experiments}

We compare \name with a variety of baselines, including methods that utilize either meshes or Gaussian Splatting, 
some of which are also task-aware. We evaluate on 
cluttered, complex scenes from the \clio~\cite{Maggio24ral-clio} dataset (\cref{sec:clio_experiments}), 
the \lerf dataset~\cite{Kerr23iccv-lerf} (\cref{sec:lerf_experiments}), and show qualitative results of using 
different tasks list on a table-top scene in \cref{fig:cover_fig}.
These experiments demonstrate \name improves upon accuracy of all methods and substantially 
reduces memory and runtime compared to other Gaussian Splatting based approaches.
We keep the same Gaussian parameters for all datasets to show that these parameters allow \name to perform well across 
multiple scenes. 

\subsection{Clio Datasets}
\label{sec:clio_experiments}

\myParagraph{Setup}
We evaluate \name on two challenging open-set object scenes (a cubicle and a small apartment) 
presented in \clio~\cite{Maggio24ral-clio}.
We use the same task lists from Clio, thus not allowing any further prompt tuning. The Cubicle 
and Apartment scenes have 18 and 26 tasks respectively such as ``clean backpacks'' and ``find spice bottles'', 
with some tasks having multiple ground truth objects. We query each method for the ground truth number of objects for 
each task. Note we omit the 
Office scene used in~\cite{Maggio24ral-clio} as the Gaussian Splat reconstruction 
is low quality and thus does not allow for fair open-set semantic evaluation of semantic Gaussian Splat based approaches. 
We keep parameters of all methods consistent for both scenes.

We use the same metrics used in~\cite{Maggio24ral-clio} which are summarized as follows: 
\begin{itemize}
  \item \textbf{Strict-osR}: We define strict open-set recall as the ratio of detections where the minimum 3D oriented bounding box of the estimated 
  object and the ground truth object contains each others' centroid.
  \item \textbf{Relaxed-osR}: We define relaxed open-set recall as the ratio of detections where the minimum 
  3D oriented bounded box of the estimated object captures 
  the centroid of the ground truth object or vice versa.
  \item \textbf{IoU}: The average Intersection over Union of the 
  minimum oriented bounding boxes for the estimated and ground truth objects.
  \item \textbf{Objs}: The number of objects in the map.
\end{itemize}
We remark that in the worst case, relaxed accuracy can be met with bounding boxes as large as the scene space and strict accuracy 
can be missed if two objects have significant overlap but fail to mutually capture both centroids. 
Thus, including both provides a clearer picture of performance.

\myParagraph{Baseline Methods}
In addition to the methods used in~\cite{Maggio24ral-clio} which 
 use meshes (with the exception of \cg~\cite{Gu24icra-conceptgraphs} which uses point clouds), 
we include two semantic Gaussian Splatting based 
methods, \sg~\cite{Guo24arxiv-semanticGS} and \ogs~\cite{Wu24neurips-opengaussian}. Methods that have some awareness 
of the task are highlighted in color in \cref{table:custom_datasets}.

\myParagraph{Quantitative Results} 
\cref{table:custom_datasets} compares \name against baselines on the Clio datasets.
On both datasets, \name consistently performs well across all metrics (first on all metrics except being 
second on Relaxed Recall for Cubicle), while \textit{none of the other baselines 
are ever first or second across all metrics}. Since \sg does not cluster Gaussians into objects but instead picks all 
Gaussians that have the same label as the best cosine score, it results in very large bounding boxes. For example, for the task 
``get notebooks'' it correctly returns Gaussians from the notebooks but also from stacks of paper on the far side of the scene resulting in 
a large bounding box and low IoU (see \cref{sec:sem_gs} for a visual example). \ogs produces cleaner objects than \sg but does not form objects at the correct granularity causing it to underperform 
 task-aware methods. \name also outperforms the task-aware baselines due to better reasoning of semantics and 
produces much better object constructions by using Gaussian Splatting as opposed to coarse meshes (additional visualizations are included in \cref{sec:qualitative_results}). 

Given a Gaussian Splat, \name takes approximately 8 minutes and 12 minutes to map  
the Cubicle and Apartment scenes respectively, which includes the time to run \fsam and CLIP. 
Before removing the task-irrelevant objects, \name had 41 and 43 objects for the Cubicle and Apartment scenes respectively, 
with each resulting in 37 objects for the final map. The removed objects oftentimes are 
large clusters of irrelevant primitives.

\begin{table}[h]\scriptsize
    \setlength{\tabcolsep}{2pt}
    \resizebox{\columnwidth}{!}{
    \begin{tabular}{cl ccccc}
    \toprule
    Scene & Method & Strict-osR$\uparrow$ & Relaxed-osR$\uparrow$ & IoU$\uparrow$ & Objs$\downarrow$\\
    \midrule
    & CG~\cite{Gu24icra-conceptgraphs} & 0.44  & 0.61  & 0.06  & 181  \\ 
    & Khronos~\cite{Schmid24rss-khronos} & 0.78 & 0.83 & 0.17 & 628 \\
    & Clio-Prim~\cite{Maggio24ral-clio} & 0.72 & 0.72 & 0.18 & 1070 \\
    & \sg~\cite{Guo24arxiv-semanticGS} & 0.17 & \underline{0.94} & 0.03 & - \\
    & \ogs~\cite{Wu24neurips-opengaussian} & 0.78 & 0.44 & 0.07 & 640 \\
    \rowcolor{blue!20} \cellcolor{white} & CG-\thres & 0.44 & 0.61 & 0.06 & 26 \\
    \rowcolor{blue!20} \cellcolor{white}& Khronos-\thres & 0.78 & 0.83 & 0.17 & 133 \\
    \rowcolor{blue!20} \cellcolor{white}& Clio-\batch~\cite{Maggio24ral-clio} & \underline{0.83} & \textbf{1.0} & 0.17 & 48\\
    \rowcolor{blue!20} \cellcolor{white}& Clio-online~\cite{Maggio24ral-clio} & \textbf{0.89} & 0.89 & \underline{0.22} & 92 \\
    \rowcolor{blue!20} \cellcolor{white}& \name & \textbf{0.89} & \underline{0.94} & \textbf{0.26} & 37 \\
    \midrule
    \multirow{-12}{*}{\rotatebox{90}{Cubicle}}
    & CG~\cite{Gu24icra-conceptgraphs}  & 0.38 & 0.62 & 0.07 & 339 \\ 
    & Khronos~\cite{Schmid24rss-khronos} & 0.45 & \underline{0.76} & 0.11 & 1093  \\
    & Clio-Prim~\cite{Maggio24ral-clio} & 0.35 & 0.59 & \underline{0.12} & 1694 \\
    & \sg~\cite{Guo24arxiv-semanticGS} & 0.00 & 0.52 & 0.00 & - \\
    & \ogs~\cite{Wu24neurips-opengaussian} & 0.31 & 0.59 & 0.06 & 640 \\
    \rowcolor{blue!20} \cellcolor{white}& CG-\thres & 0.21 & 0.35 & 0.03 & 21 \\
    \rowcolor{blue!20} \cellcolor{white}& Khronos-\thres & 0.41 & 0.72 & 0.11 & 162 \\
    \rowcolor{blue!20} \cellcolor{white}& Clio-\batch~\cite{Maggio24ral-clio} & \underline{0.52} & 0.72 & 0.11 & 90\\
    \rowcolor{blue!20} \cellcolor{white}& Clio-online~\cite{Maggio24ral-clio} & 0.35 & 0.52 & 0.07 & 99 \\
    \rowcolor{blue!20} \cellcolor{white}& \name & \textbf{0.59} & \textbf{0.90} & \textbf{0.19} & 37 \\
    \midrule
    \multirow{-12}{*}{\rotatebox{90}{Apartment}}
    \end{tabular}
    }
    \vspace{-4mm}
    \caption{Results from the Cubicle and Apartment scenes from the \clio dataset~\cite{Maggio24ral-clio}. 
    Highlighted methods have some awareness of the tasks during mapping. Number of objects is omitted for 
    \sg since it does not cluster Gaussians together. Number of objects for \ogs is given as the number of leaves in the 
    fine codebook. \textbf{best}, \underline{second best}
    } 
    \label{table:custom_datasets}
\end{table} 
\myParagraph{Memory Usage}
We demonstrate the memory advantage of \name in \cref{table:memory}. 
Since we do not attach CLIP vectors to Gaussians, we use an order of magnitude less 
memory compared to \sg when computing $p(y|\phi)$ for all Gaussians. As we compress the scene 
into primitives and finally into task-relevant objects, the memory further decreases.
Importantly, each stage of clustering (Gaussians, 3D primitives, and task-relevant objects) 
{does not require using memory from the other stages} (i.e., we only need about 3 Mb for our final map). We include a breakdown of the memory usage of each stage of 
\name in \cref{table:memory_breakdown}. 
The memory of each stage consists of attaching probabilities to each item in the stage. Additionally, we have IDs on each 
Gaussian to associate them to primitives. While this is not needed for 
the Gaussian stage, for efficiency it is computed here 
since both computing semantic probabilities for each Gaussian and assignment to primitives require segmentation masks. 
Finally, the objects require assignment mapping of primitive IDs to objects.
The substantial memory reduction further underscores the advantage of creating task-relevant object maps.
Memory can be further reduced by removing the task-irrelevant Gaussians, but currently we keep these available in case 
downstream operations such as getting scene geometry for path planning require them.

\begin{table}[h]
  \centering
  \begin{tabular}{lccc}
    \toprule
     & Cubicle & Apartment \\
    \midrule
    GS no semantics & 274  & 294 \\
    \midrule
    \sg & 1616  & 1739 \\
    \ogs & 20.9  & 21.4 \\
    Ours All Gaussians & 15.2  & 22.8 \\
    Ours All Primitives & \underline{2.89}  & \underline{3.3} \\
    Ours All Objects & \textbf{2.77}  & \textbf{3.0} \\
    \bottomrule
  \end{tabular}
  \caption{Memory usage in Mb for the Cubicle and Apartment scenes. 
  Memory for each method is listed and only captures the storage of semantic information (\ie not including attributes such as RGB and covariance
  which are included as the first entry).
  \textbf{We include memory for \name at three stages of clustering.} 
  Note that \ogs uses a different Gaussian Splat training which 
  has higher memory for \textit{GS no semantics}, and is not included here.}
  \label{table:memory}
  \vspace{-5pt}
\end{table}

  \begin{table}[h]
    \centering
    \begin{tabular}{lccc}
      \toprule
      Layer & Gaussians & Primitives & Objects \\
      \midrule
      Total Memory & 23  & 3.26 & 3.05 \\
      \midrule
      Probabilities & 19.75  & 0.22 & 0.002 \\
      Gaussian IDs & 3.04  & 3.04 & 3.04 \\
      Cluster IDs & N/A  & N/A & 0.0001 \\
      \bottomrule
    \end{tabular}
    \caption{Memory usage (Mb) breakdown of each component of the stages of \name on the Apartment dataset. 
    {Each stage does not require using memory from the other stages (\ie the resulting map needs only 3.05 Mb for semantics.)}
    There are about 400,000 Gaussians, 1108 primitives, and 37 objects.}
    \label{table:memory_breakdown}
    \vspace{-5pt}
  \end{table}

\subsection{\lerf Dataset Results}
\label{sec:lerf_experiments}

We also provide results of running \name on the small-scale LeRF dataset~\cite{Kerr23iccv-lerf} using the evaluation protocol 
of \ogs~\cite{Wu24neurips-opengaussian}. We keep all parameters of \name the same as the ones in \cref{sec:clio_experiments} to demonstrate 
the stability of our method. 
We query \name for the set of 3D objects for each ground truth text label and compute IoU and Accuracy which 
here is the percent of objects with an IoU 
greater than 0.25. The set is defined as all the objects which have a probability 
of at least 0.8 that of the most similar object, which follows \ogs's protocol
of using a fraction of the top cosine score.
\name uses the list of text labels as its task list during 
map construction.
\name either performs the best or is second to \ogs despite running substantially faster and has 
the highest average score for both IoU and Accuracy. 
\ogs uses CLIP and SAM embeddings from LangSplat~\cite{Qin24cvpr-langsplat} which uses approximately 1 hour and 
then averages 50 minutes 
to run semantic optimization given a trained 
Gaussian Splat, while \name takes 4 minutes to run all steps given a 
trained Gaussian Splat, which includes the time to run \fsam and CLIP.

\begin{table}[h]
\centering
\resizebox{\columnwidth}{!}{
\begin{tabular}{c|c|c|c|c|c}
\toprule
\multirow{2}{*}{Methods} & \texttt{figurines} & \texttt{teatime} & \texttt{ramen} & \texttt{kitchen} & \textbf{Mean} \\
                         & mIoU/mAcc         & mIoU/mAcc        & mIoU/mAcc       & mIoU/mAcc             & mIoU/mAcc \\
\midrule
LangSplat~\cite{Qin24cvpr-langsplat}   & 10.16 / 8.93  & 11.38 / 20.34  & 7.92 / 11.27  & 9.18 / 9.09  & 9.66 / 12.41 \\
LEGaussians~\cite{Shi24cvpr-LEGaussians} & 17.99 / 23.21 & 19.27 / 27.12 & 15.79 / 26.76 & 11.78 / 18.18 & 16.21 / 23.82 \\
OpenGaussian~\cite{Wu24neurips-opengaussian} & \underline{39.29} / \underline{55.36} & \textbf{60.44} / \textbf{76.27} & \textbf{31.01} / \underline{42.25} & \underline{22.70} / \underline{31.82} & \underline{38.36} / \underline{51.43} \\
\name (ours) & \textbf{46.70} / \textbf{67.86} & \underline{52.76} / \underline{71.19} & \underline{29.12} / \textbf{45.07} & \textbf{28.57} / \textbf{54.55} & \textbf{39.29} / \textbf{59.67} \\
\bottomrule
\end{tabular}
}
\caption{Results of average IoU and Accuracy on four scenes from the LeRF dataset~\cite{Kerr23iccv-lerf}. 
\textbf{best}, \underline{second best}}
\label{tab:lerf}
\vspace{-2mm}
\end{table}

\section{Limitations}
\label{sec:limitations}

While \name improves over baselines for semantic mapping, it
is not without limitations. Firstly, we still inherit some limitations from the underlying foundation model, such as being affected 
by the choice of prompts.
Additionally, the prompts must be specific concepts such as ``get dish soap'' and not general tasks 
such as ``items needed for a cleaning robot.'' 
Due to the limited receptive field, it is also possible that relevant parts 
of an object are not correctly merged if the parts by themselves have little relevance to the task. 
Similarly, if we observe a relevant object 
though multiple partial or obstructed views that all have low probability of being task-relevant, then we 
are unable to reason that a task-relevant object was observed.

\section{Conclusion}
\label{sec:conclusion}

We have presented \name, a method which performs task-driven semantic mapping with Gaussian Splatting and allows for 
extraction of 3D objects at the correct task-determined granularity. 
By developing a probabilistic measurement model to describe the relationship between visual 
observations and natural language, \name is able to use Bayesian updating to aggregate semantic information 
across multiple observations of the scene. In addition to improving upon accuracy of object mapping, our method's task-driven 
structure greatly reduces memory usage. Lastly, given a vanilla Gaussian Splat, 
our method runs in minutes without requiring any training or scene-specific network weights, making it 
suitable for future work as an incremental 
real-time mapping pipeline.
\section*{Acknowledgments}
This work is supported in part 
by the NSF Graduate Research Fellowship Program under Grant 2141064, 
{and the ONR RAPID program.}
The authors gratefully acknowledge Yun Chang for thoughtful discussions about the Information Bottleneck. {
    \small
    \bibliographystyle{ieeenat_fullname}

}

\clearpage
\setcounter{page}{1}
\maketitlesupplementary

\appendix
\renewcommand{\appendixname}{}  %

\section{Ablations}
\label{sec:ablations}
To demonstrate the effectiveness of our probabilistic inference model and components of \name, 
we ablate multiple configurations in \cref{table:ablations}. 
Here we show Strict-OsR, Relaxed-OsR, and IoU results averaged across the \clio Apartment and Cubicle datasets for the following studies:
\begin{itemize}
    \item \textbf{OR-$\phi$}: Using our outlier rejection protocol for cosine scores based on our measurement model.
    \item \textbf{kNN}: Using k-nearest neighbor point cloud outlier rejection for 3D objects.
    \item \textbf{BU}: Performing Bayesian updating. The alternative is to simple average cosine scores across all
    views and get $p(y|x)$ for the \aib using \cref{eq:bayes} with the averaged cosine scores as input. 
    This is equivalent to averaging CLIP vectors across views assuming the vectors are first normalized. 
    Note that we still can run outlier rejection (OR-$\phi$) by also computing the probability. 
    \item \textbf{PE}: Extracting the most relevant objects for each task as described in \cref{sec:get_objects} with the probability 
    to each task instead of computing a 
    CLIP vector based on rendering of the task-relevant object and selecting the object 
    with the highest cosine score to each task.
\end{itemize}

\begin{table}[h]
  \renewcommand\tabcolsep{6pt}  %
  \renewcommand{\arraystretch}{1.1}  %
  \centering
  \begin{tabular}{cccc|ccc}
    \toprule
    OR-$\phi$ & kNN & BU & PE & S-OsR  & R-OsR &  IoU\\
    \midrule
       & \Checkmark & \Checkmark & \Checkmark & 0.33 & \textbf{0.94} & 0.09 \\
      \Checkmark & & \Checkmark & \Checkmark & 0.51 & 0.81 & 0.14 \\
      \Checkmark & \Checkmark & & \Checkmark & \underline{0.54} & 0.81 & \underline{0.18}\\
      \Checkmark & \Checkmark & \Checkmark &  & 0.42 & 0.54 & 0.13 \\
      \Checkmark & \Checkmark & \Checkmark & \Checkmark & \textbf{0.74} & \underline{0.92} & \textbf{0.22}\\
    \bottomrule
  \end{tabular}
  \caption{Ablation studies of performance of \name with results shown as averages for the Apartment and Cubicle datasets.}
  \label{table:ablations}
  \vspace{-8pt}
\end{table}

We see that our configuration performs the best across all ablations and all metrics except on Relaxed Recall where it is 
second to the first configuration which does not use our probabilistic outlier rejection (OR-$\phi$). 
However, the first configuration performs substantially worse on IoU and Strict-OsR since it frequently over clusters. 
This is an example of how a high score can be reached on relaxed accuracy by over clustering and highlights the benefit of 
including both accuracy metrics.

Using k-nearest neighbor improves performance substantially, especially on IoU and Strict-OsR as 
a few stray Gaussians resulting from either noisy SAM masks or imperfect scene reconstruction can affect the 
estimated minimum 3D bounding box of the objects.
We also importantly observe that using our Bayesian updating approach (BU) for aggregating information across multiple views performs substantially better 
than simply averaging cosine scores or averaging CLIP vector. We also see that our default configuration of selecting objects based on the highest 
probability to the tasks (PE) performs better than computing the CLIP vector for each object due to robustness of considering semantic 
information across multiple observations.

\section{Extra Visual Results}
\label{sec:qualitative_results}

We include additional visual results across the Clio and LeRF datasets. Each shows a Gaussian 
Splatting RGB rendering along with a rendering of semantic labels where each object is assigned a different color. 
We also show extraction of select objects since our system is designed for 3D object extraction. Each of the shown extracted 
objects is the most relevant object for one of the tasks in the task lists (\cref{eq:select_object}).
Objects that have been removed by \name for being designated not task-relevant 
are colored white. Here we can observe that the objects estimated as being not task-relevant 
are often made up of large clusters of Gaussians, as significant portions of the scenes are colored white. 

As our primary objective is to form objects at the correct granularity to support relevant tasks, 
we do not attempt to impose strict pruning of objects that are likely irrelevant (\cref{sec:get_objects}) and thus 
some objects can remain in the final map which do not have a high probability of being relevant for the tasks. For example, 
in \cref{fig:apartment_visuals} which shows images from the Apartment scene, the semantic object instance colored purple in all views 
corresponds to the same large object which consists of Gaussians throughout the scene. This object is correctly never chosen 
as the most relevant object for any task, but since it was not pruned, it remains in the final map.

\newcommand{\hval}{4.7}

\begin{figure*}[h]
    \centering
    
    \begin{minipage}[t]{\textwidth}
        \includegraphics[width=\linewidth]{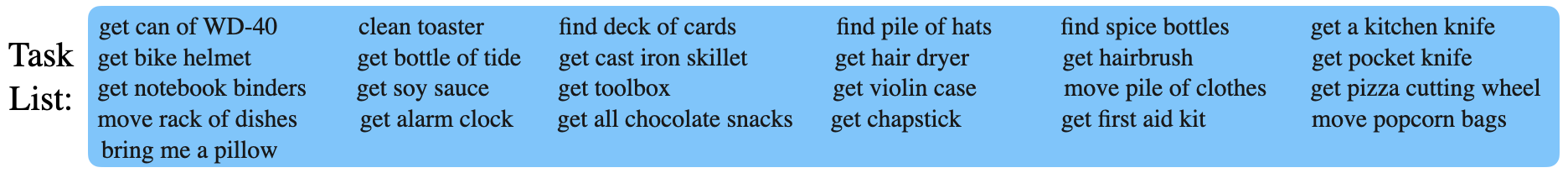}
    \end{minipage}
    
    \vspace{0.5em}
    \begin{minipage}[t]{0.49\textwidth}
      \includegraphics[width=\linewidth,height=\hval cm]{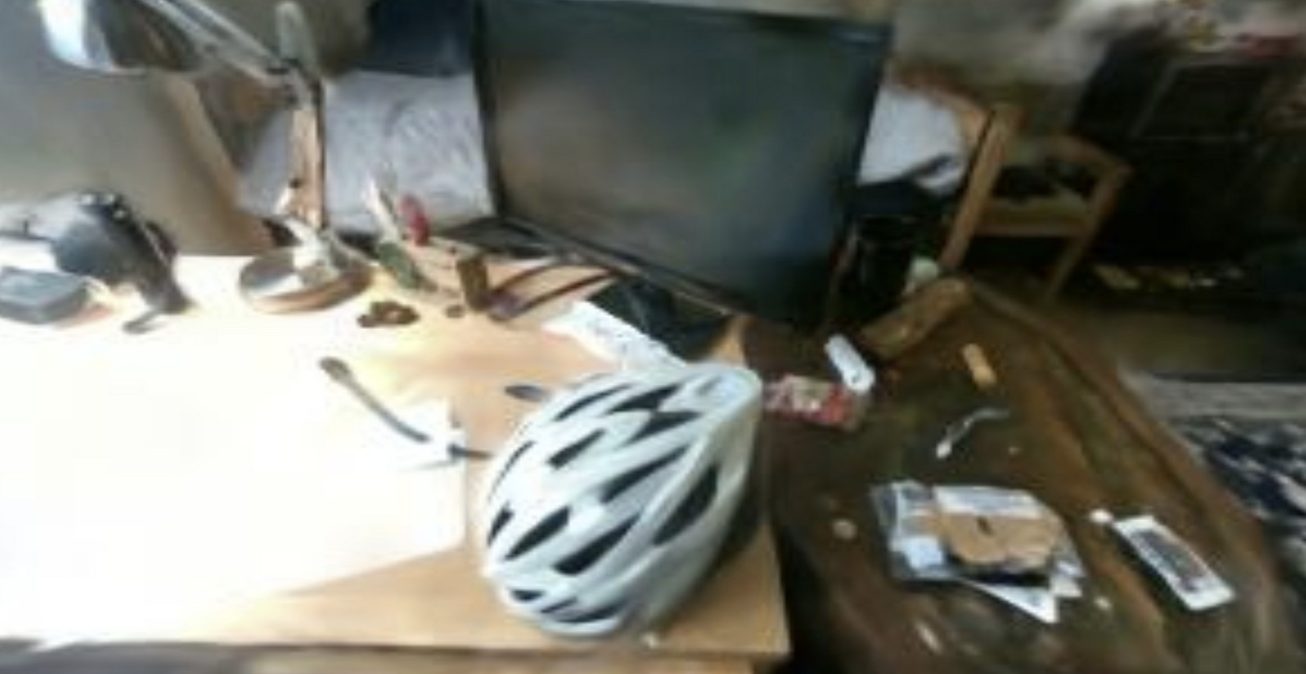}
    \end{minipage}
    \hfill
    \begin{minipage}[t]{0.49\textwidth}
      \centering
      \includegraphics[width=.8\linewidth,height=\hval cm]{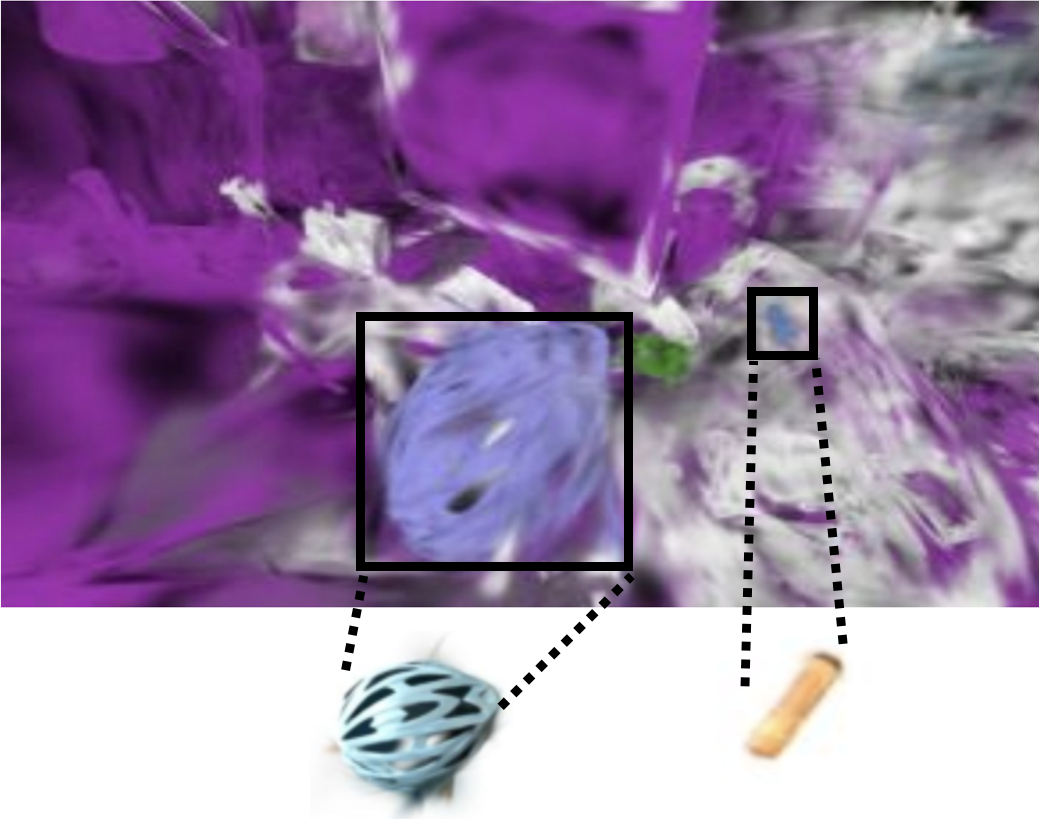}
    \end{minipage}
  
    \vspace{0.5em}
    \begin{minipage}[t]{0.49\textwidth}
      \includegraphics[width=\linewidth,height=\hval cm]{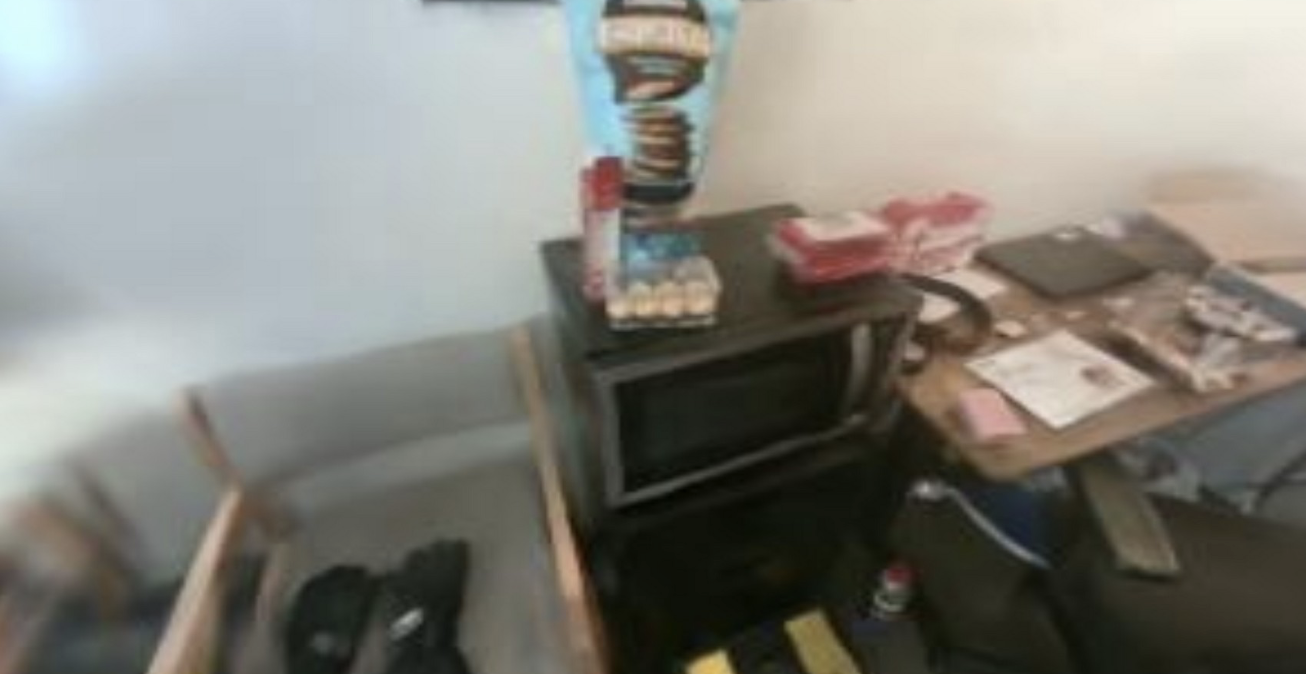}
    \end{minipage}
    \hfill
    \begin{minipage}[t]{0.49\textwidth}
      \centering
      \includegraphics[width=.8\linewidth,height=\hval cm]{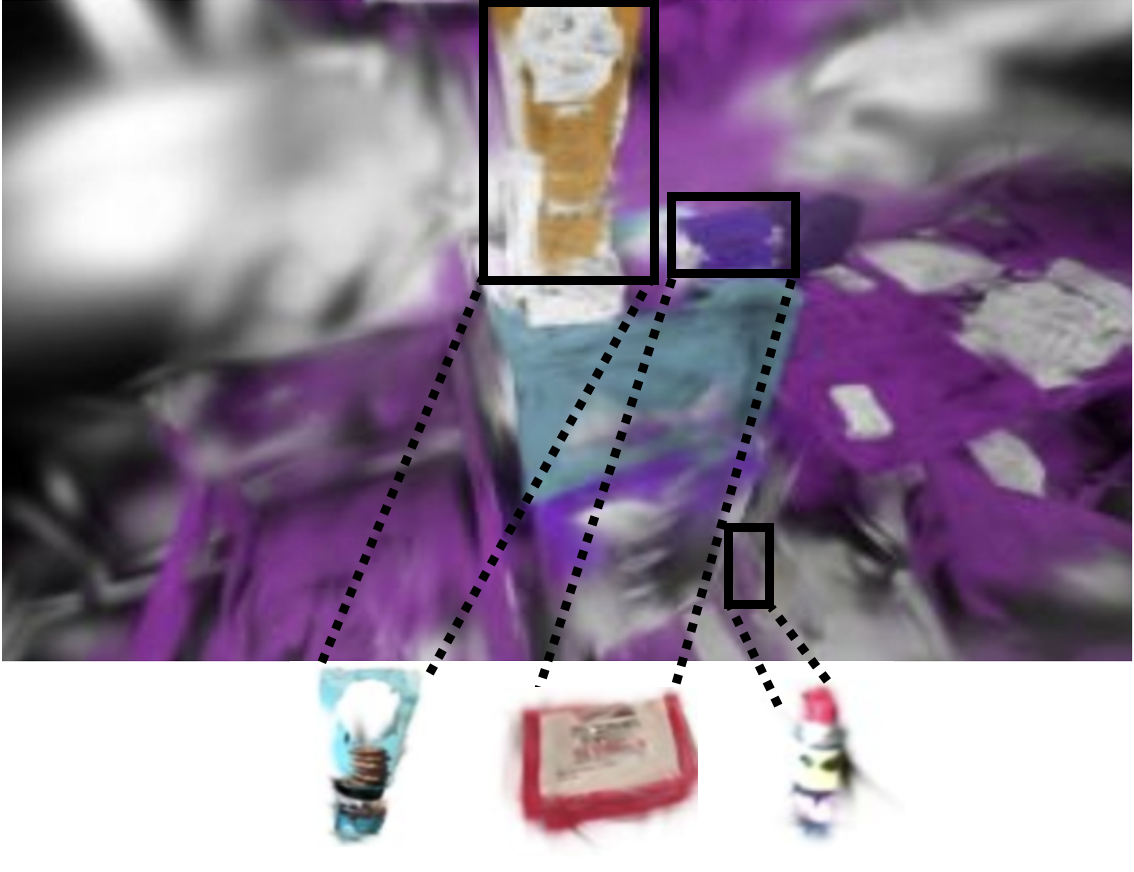}
    \end{minipage}

    \vspace{0.5em}
    \begin{minipage}[t]{0.49\textwidth}
      \includegraphics[width=\linewidth,height=\hval cm]{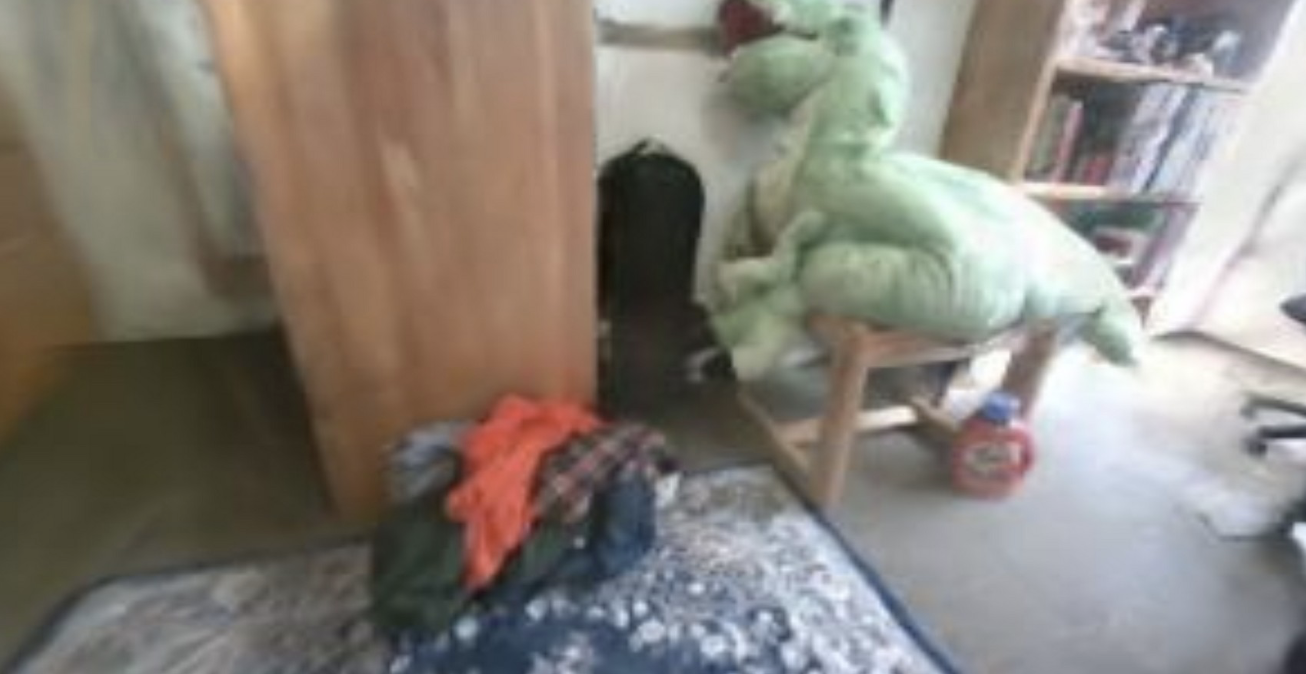}
    \end{minipage}
    \hfill
    \begin{minipage}[t]{0.49\textwidth}
      \centering
      \includegraphics[width=.8\linewidth,height=\hval cm]{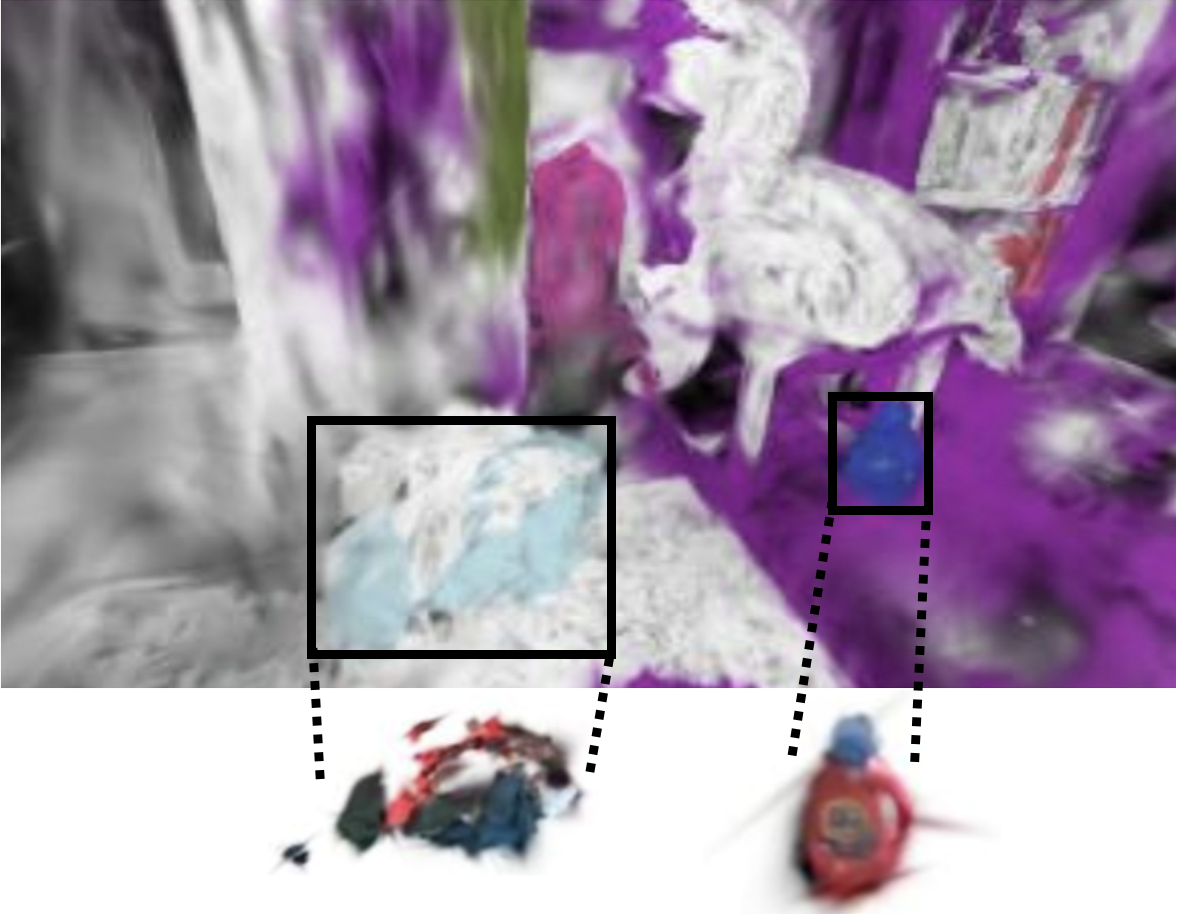}
    \end{minipage}

    \vspace{0.5em}
    \begin{minipage}[t]{0.49\textwidth}
      \includegraphics[width=\linewidth,height=\hval cm]{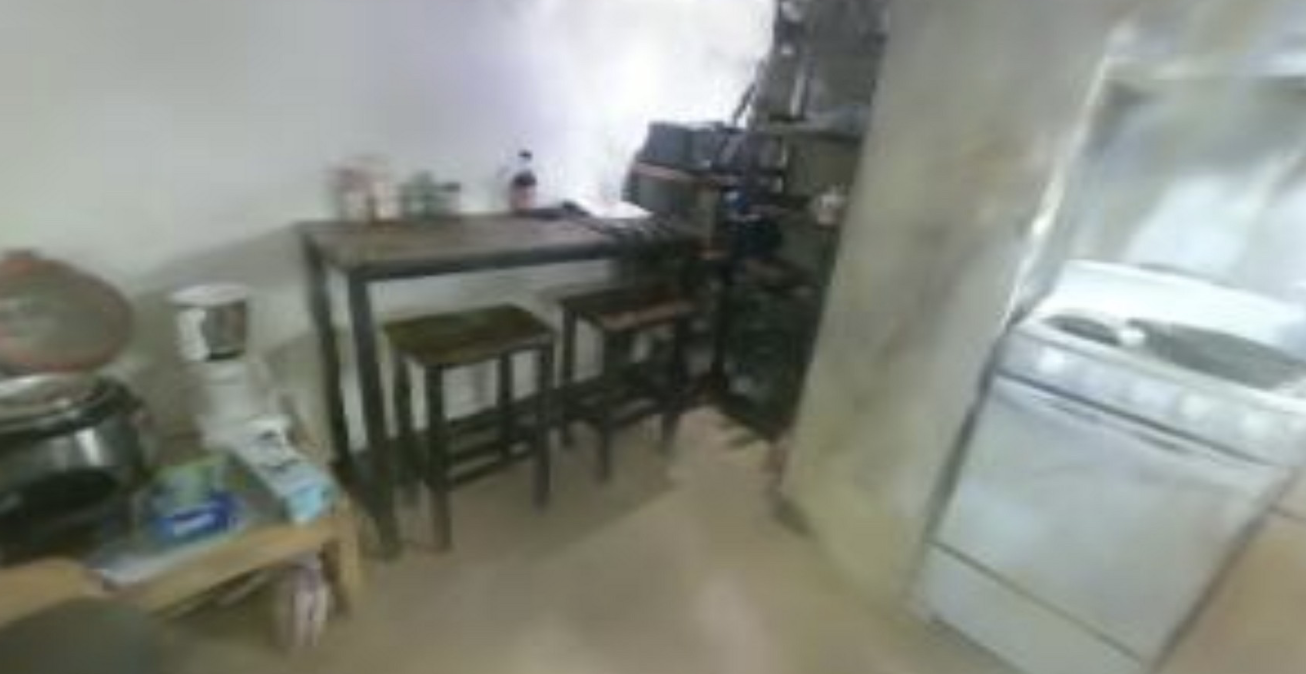}
    \end{minipage}
    \hfill
    \begin{minipage}[t]{0.49\textwidth}
      \centering
      \includegraphics[width=.8\linewidth,height=\hval cm]{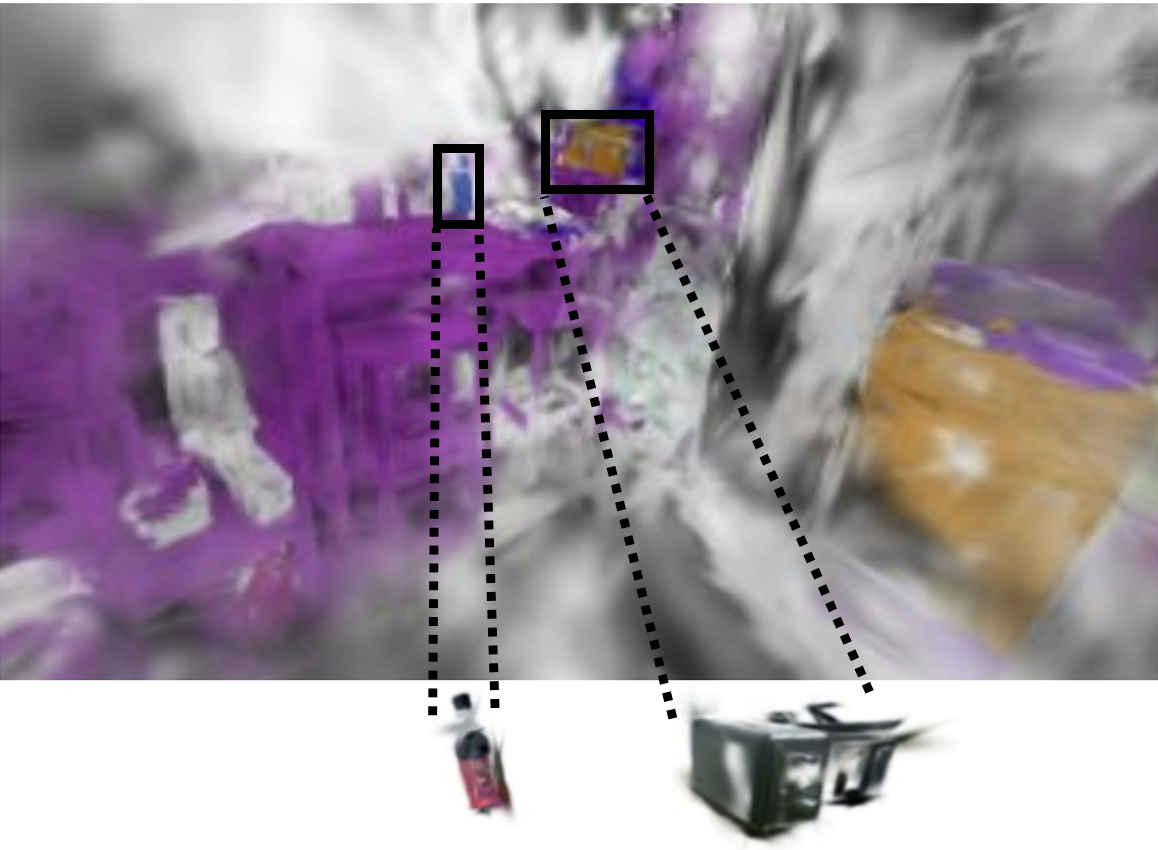}
    \end{minipage}
  
    \caption{Clio Apartment dataset results of final task-relevant objects. Left: RGB rendering, Right: semantic rendering with 
    select extracted 3D objects}
    \label{fig:apartment_visuals}
\end{figure*}

\begin{figure*}[h]
  \centering

  \begin{minipage}[t]{\textwidth}
      \includegraphics[width=\linewidth]{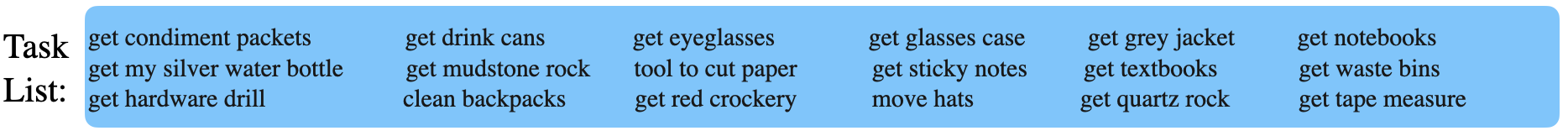}
  \end{minipage}
  
  \vspace{0.5em}

  \begin{minipage}[t]{0.49\textwidth}
    \includegraphics[width=\linewidth,height=\hval cm]{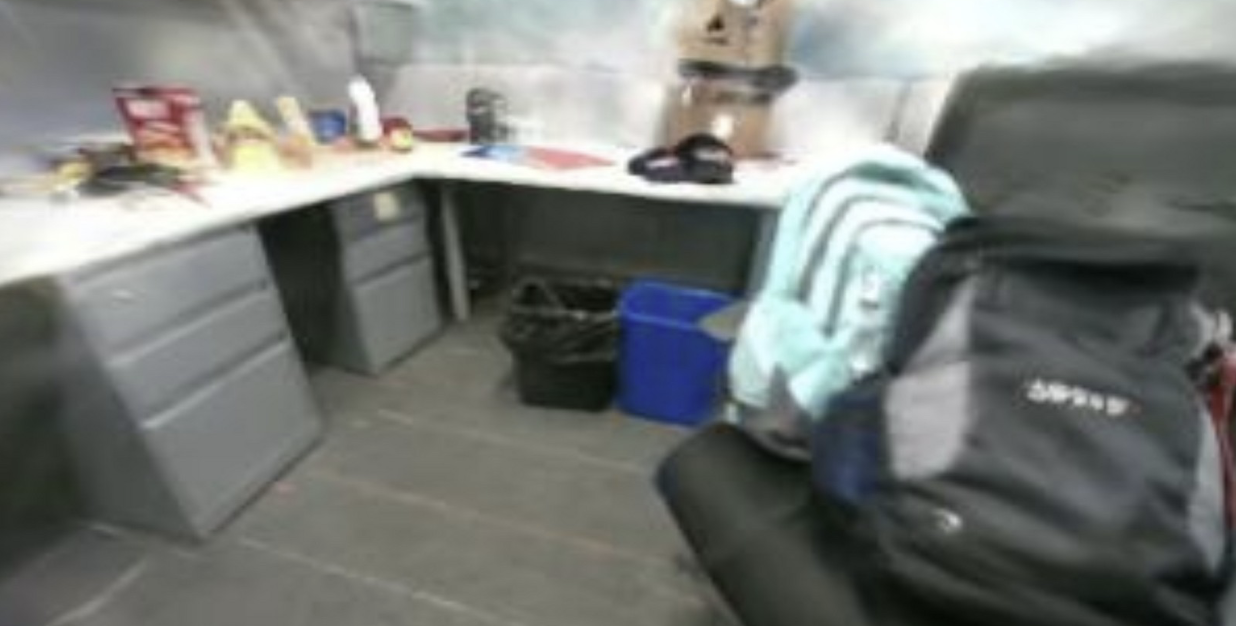}
  \end{minipage}
  \hfill
  \begin{minipage}[t]{0.49\textwidth}
    \centering
    \includegraphics[width=.8\linewidth,height=\hval cm]{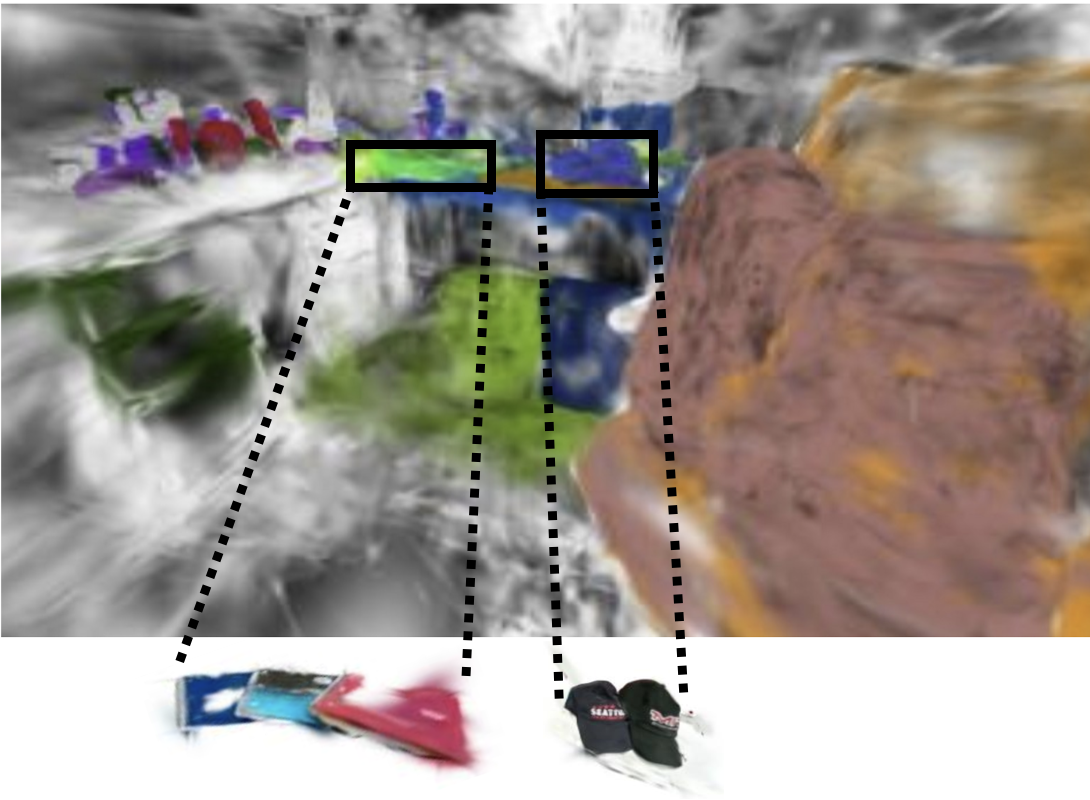}
  \end{minipage}

  \vspace{1em}

  \begin{minipage}[t]{0.49\textwidth}
    \includegraphics[width=\linewidth,height=\hval cm]{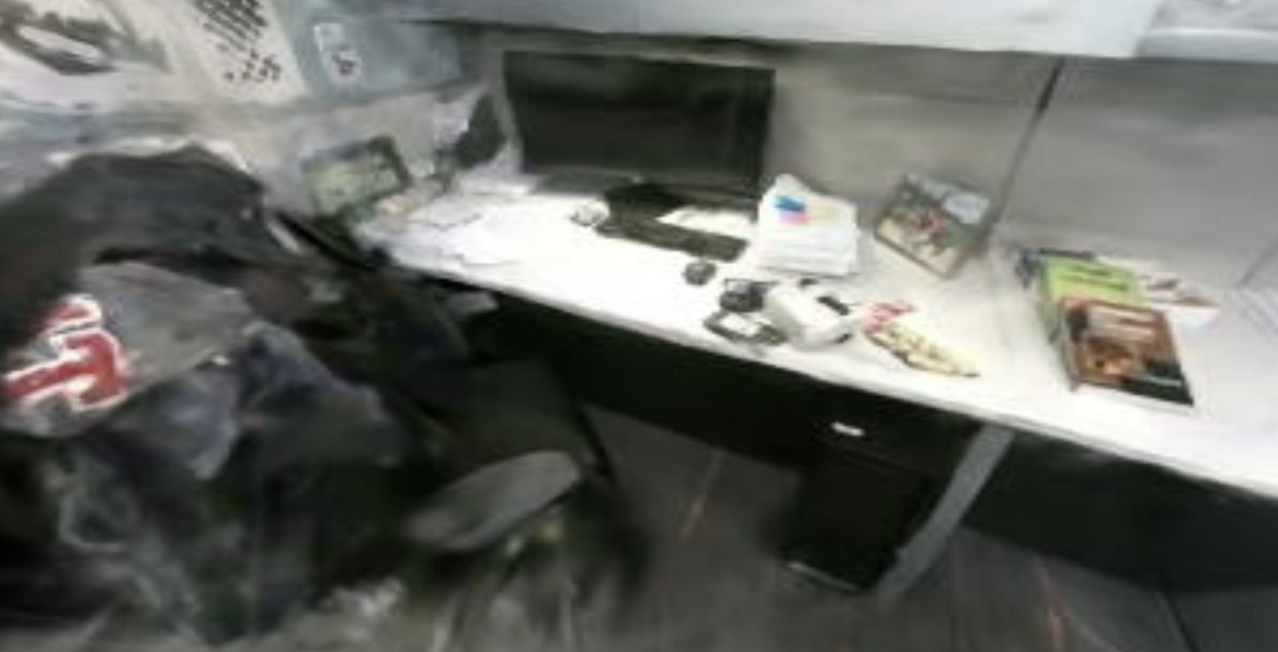}
  \end{minipage}
  \hfill
  \begin{minipage}[t]{0.49\textwidth}
    \centering
    \includegraphics[width=.8\linewidth,height=\hval cm]{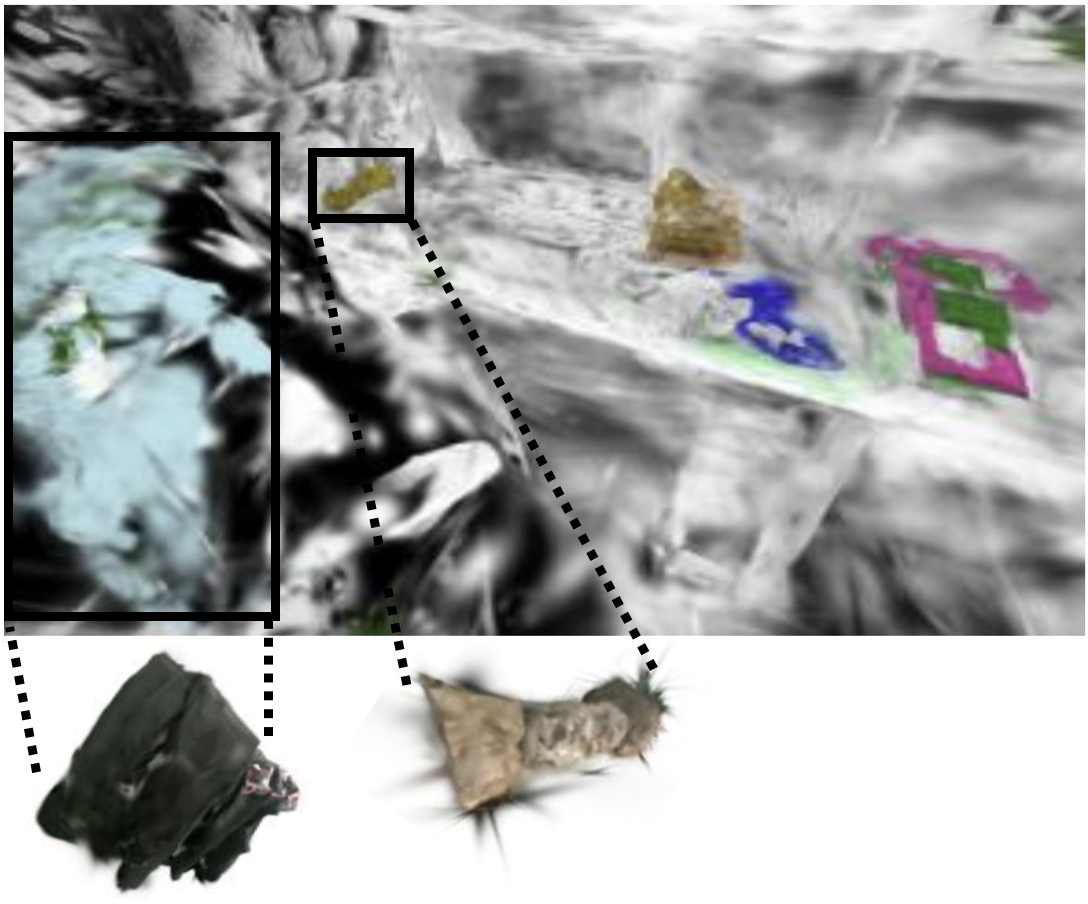}
  \end{minipage}

  \caption{Clio Cubicle dataset results of final task-relevant objects. Left: RGB rendering, Right: semantic rendering with 
  select extracted 3D objects}
  \label{fig:cubicle_visuals}
\end{figure*}

\begin{figure*}[h]
    \centering
    \begin{minipage}[t]{\textwidth}
      \includegraphics[width=\linewidth]{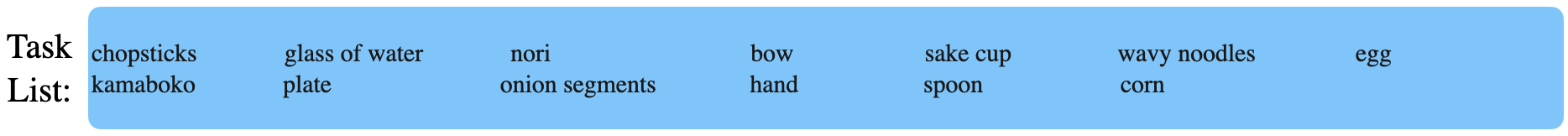}
    \end{minipage}
    
    \vspace{1em}
    \begin{minipage}[t]{0.49\textwidth}
      \includegraphics[width=\linewidth,height=\hval cm]{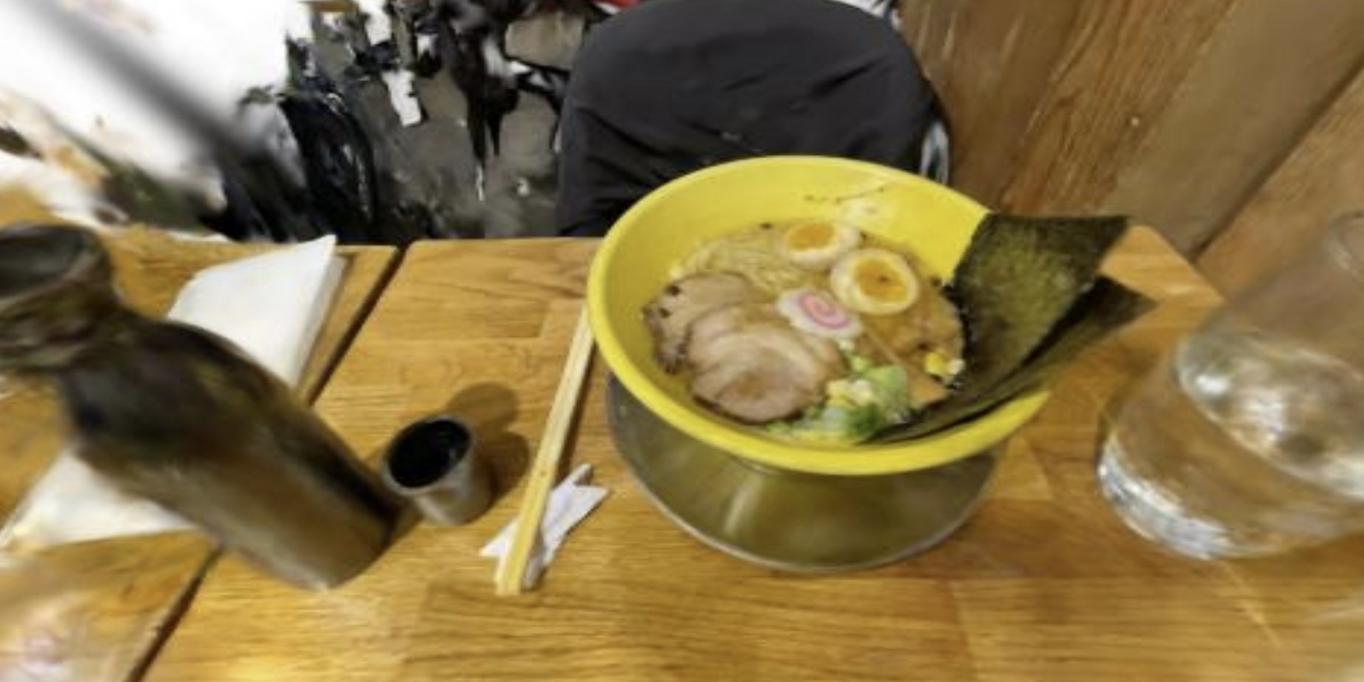}
    \end{minipage}
    \hfill
    \begin{minipage}[t]{0.49\textwidth}
      \centering
      \includegraphics[width=.8\linewidth,height=\hval cm]{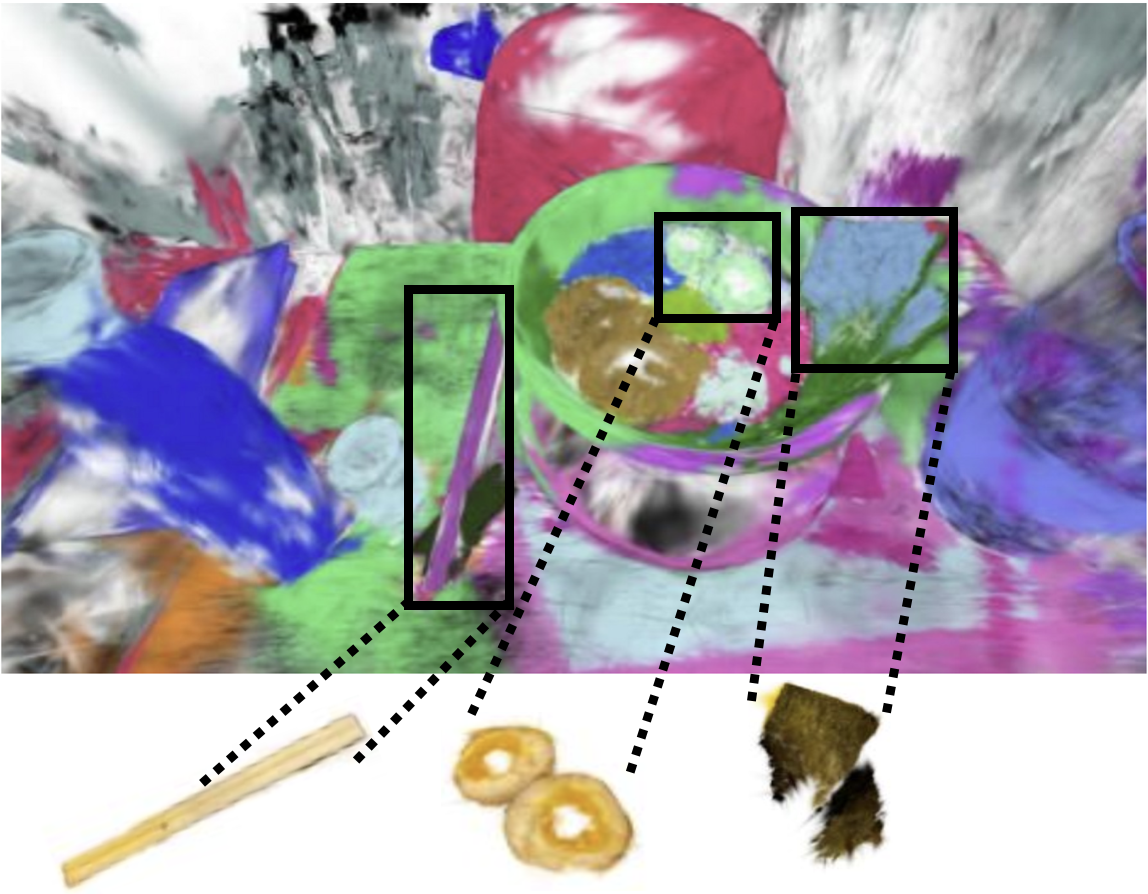}
    \end{minipage}
    \caption{LeRF Ramen dataset results of final task-relevant objects. Left: RGB rendering, Right: semantic rendering with 
    select extracted 3D objects}
    \label{fig:ramen_visuals}
\end{figure*}

\begin{figure*}[h]
    \centering
    \begin{minipage}[t]{\textwidth}
        \includegraphics[width=\linewidth]{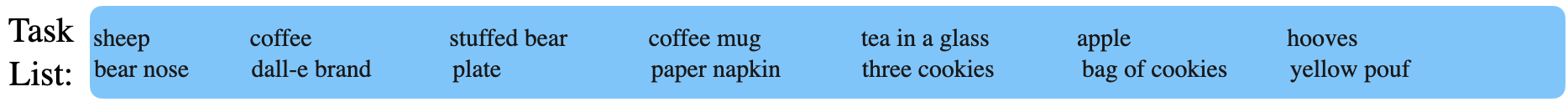}
    \end{minipage}
    
    \vspace{1em}
    \begin{minipage}[t]{0.49\textwidth}
      \includegraphics[width=\linewidth,height=\hval cm]{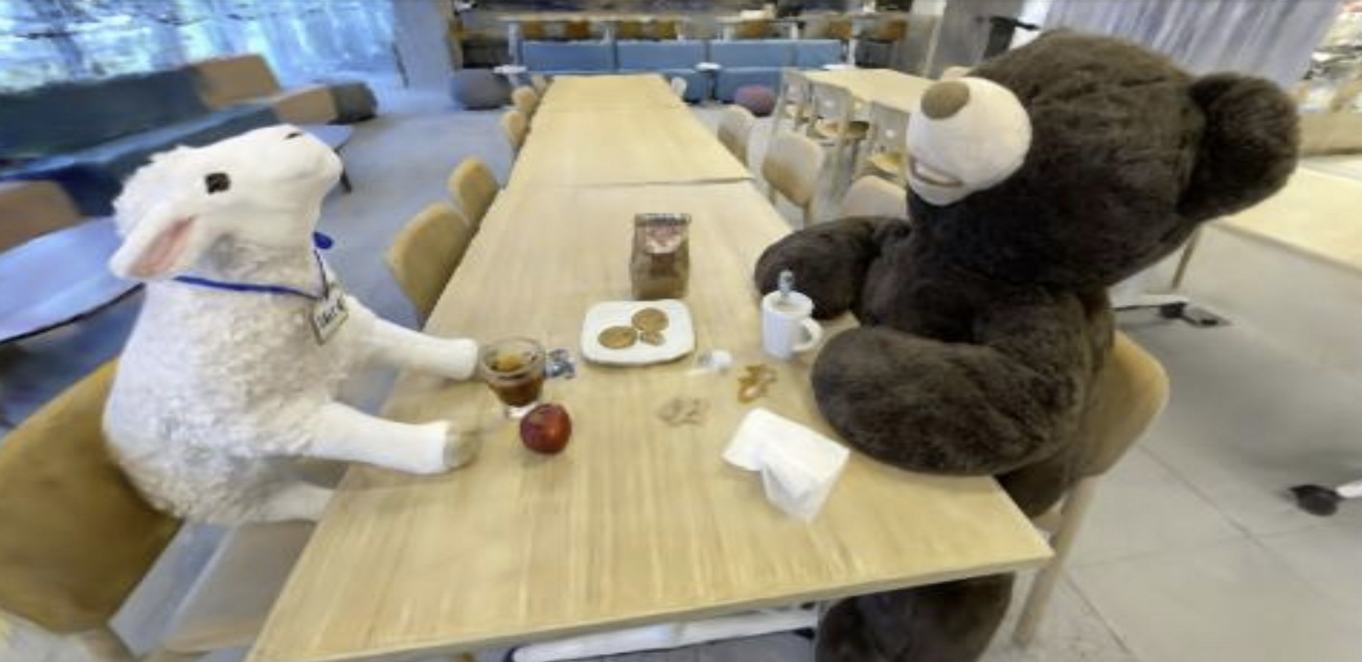}
    \end{minipage}
    \hfill
    \begin{minipage}[t]{0.49\textwidth}
      \centering
      \includegraphics[width=.8\linewidth,height=\hval cm]{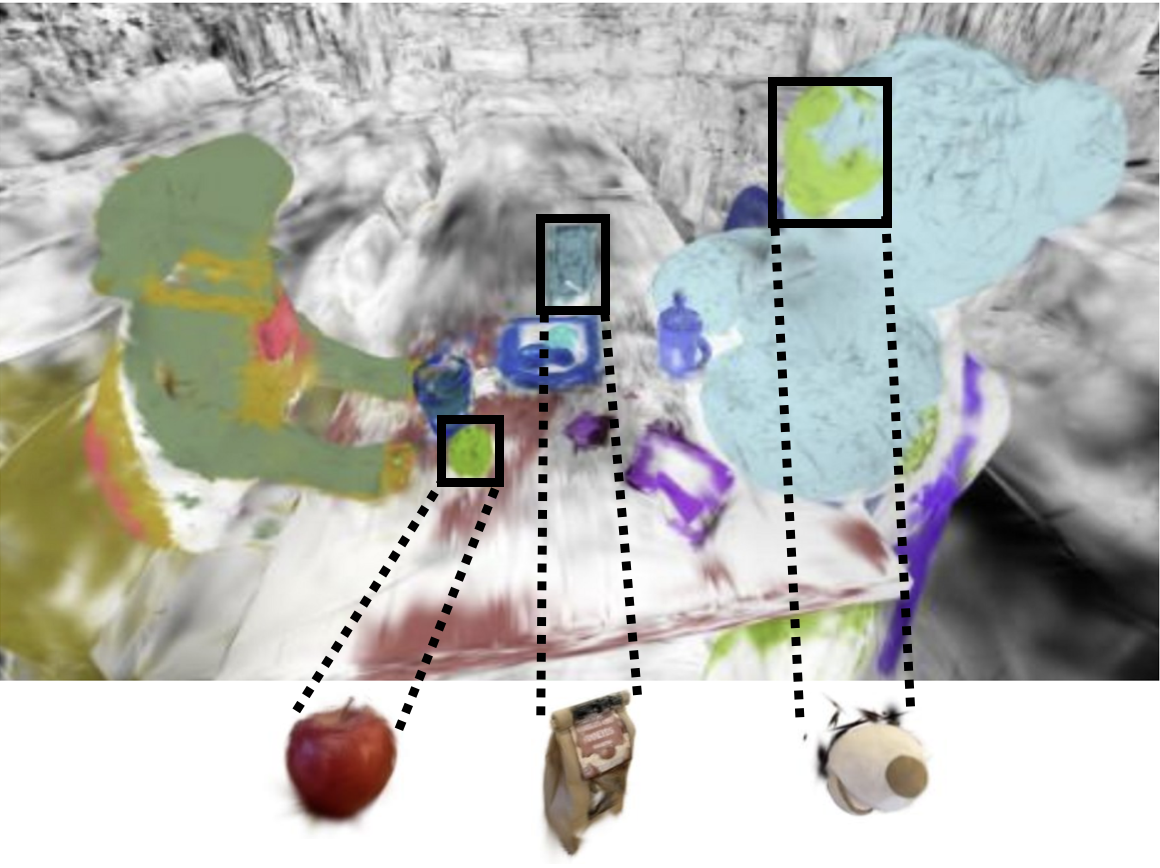}
    \end{minipage}
    \caption{LeRF Teatime dataset results of final task-relevant objects. Left: RGB rendering, Right: semantic rendering with 
    select extracted 3D objects}
    \label{fig:teatime_visuals}
\end{figure*}

\begin{figure*}[h]
    \centering
    \begin{minipage}[t]{\textwidth}
        \includegraphics[width=\linewidth]{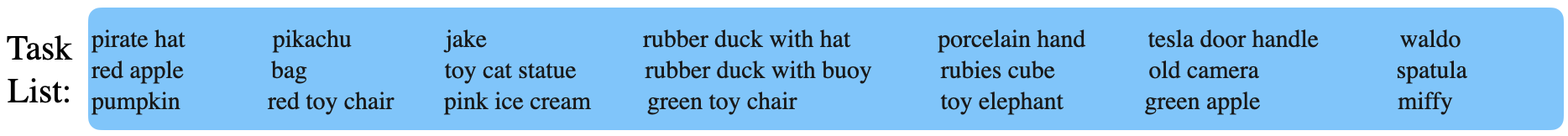}
    \end{minipage}
    
    \vspace{1em}
    \begin{minipage}[t]{0.49\textwidth}
      \includegraphics[width=\linewidth,height=\hval cm]{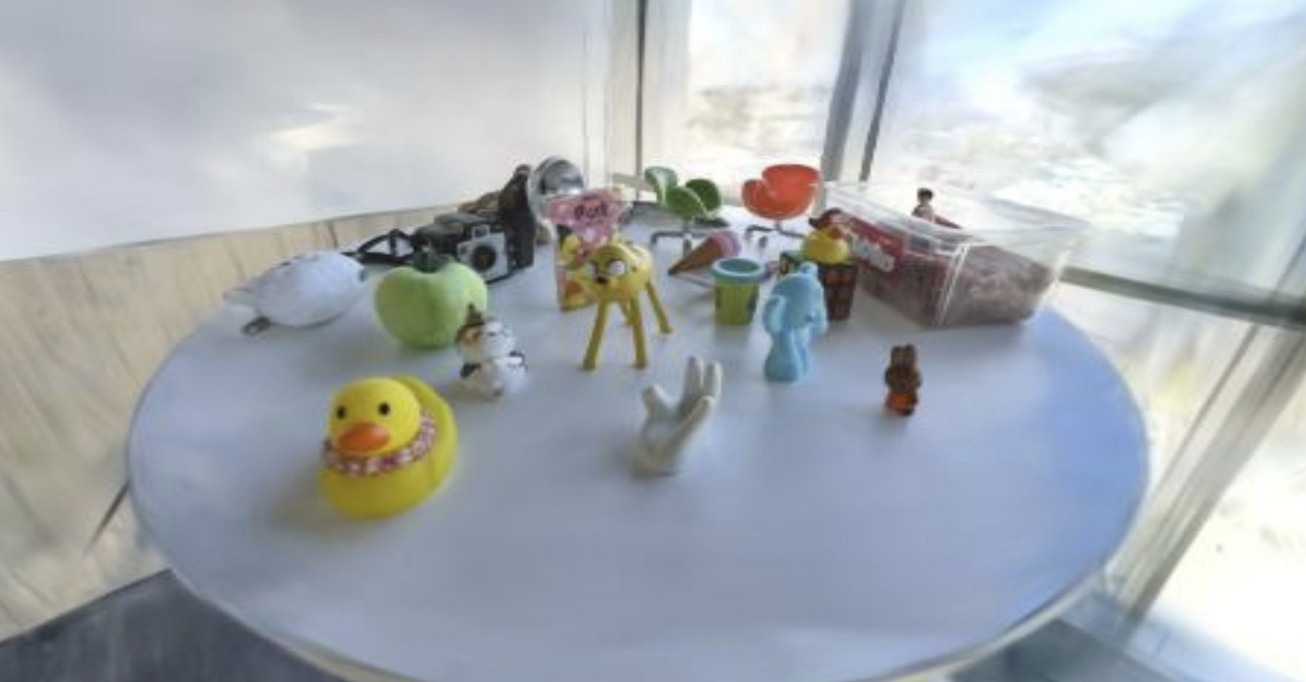}
    \end{minipage}
    \hfill
    \begin{minipage}[t]{0.49\textwidth}
      \centering
      \includegraphics[width=.8\linewidth,height=\hval cm]{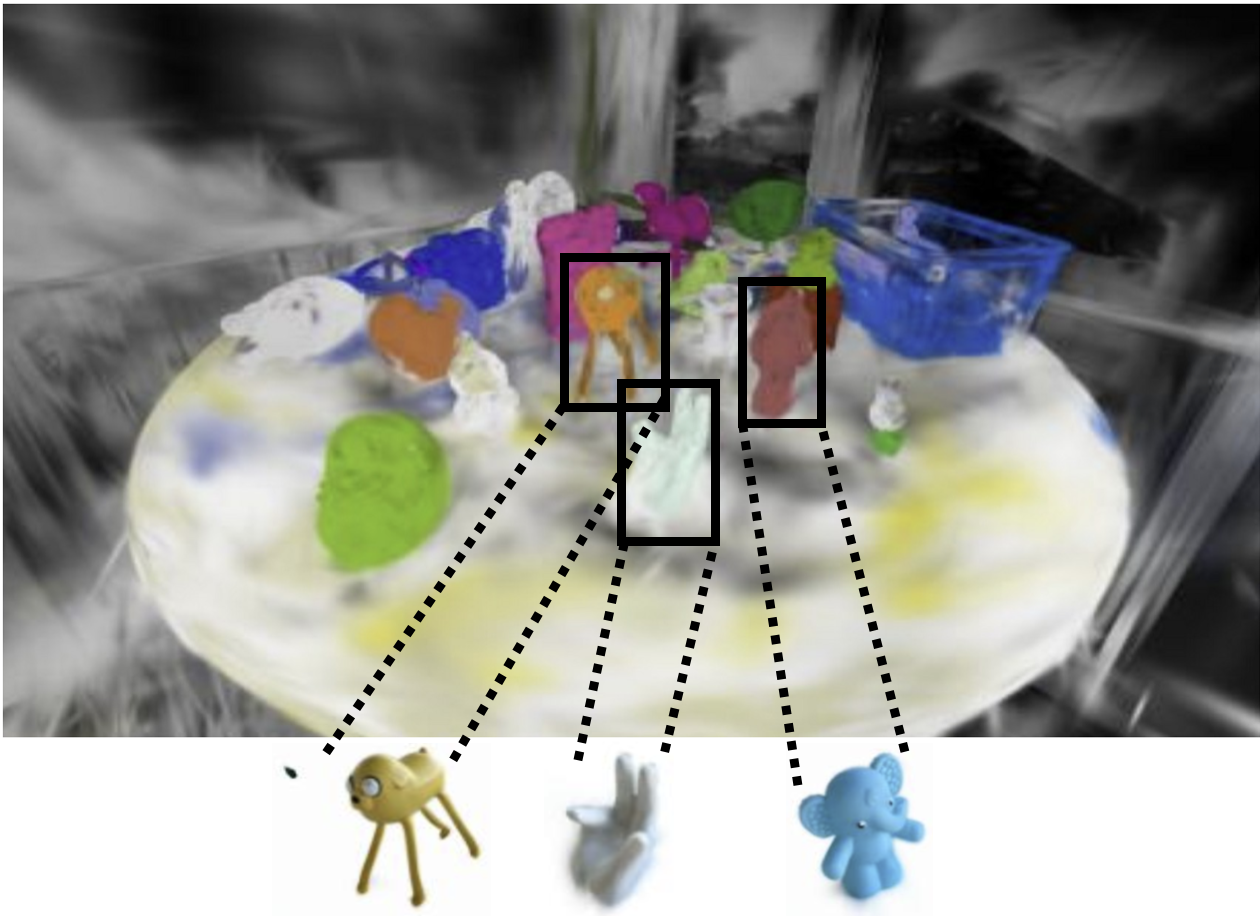}
    \end{minipage}
    \caption{LeRF Figurines dataset results of final task-relevant objects. Left: RGB rendering, Right: semantic rendering with 
    select extracted 3D objects}
    \label{fig:figurines_visuals}
\end{figure*}

\begin{figure*}[h]
    \centering
    \begin{minipage}[t]{\textwidth}
        \includegraphics[width=\linewidth]{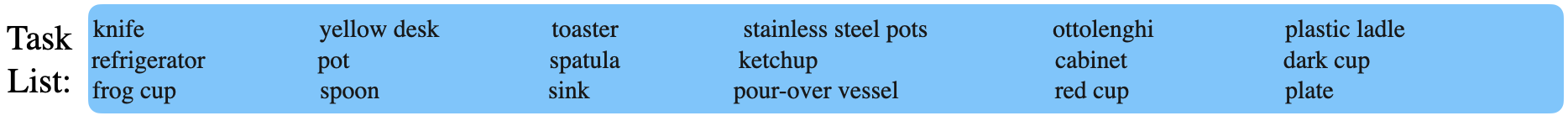}
    \end{minipage}
    
    \vspace{1em}
    \begin{minipage}[t]{0.49\textwidth}
      \includegraphics[width=\linewidth,height=\hval cm]{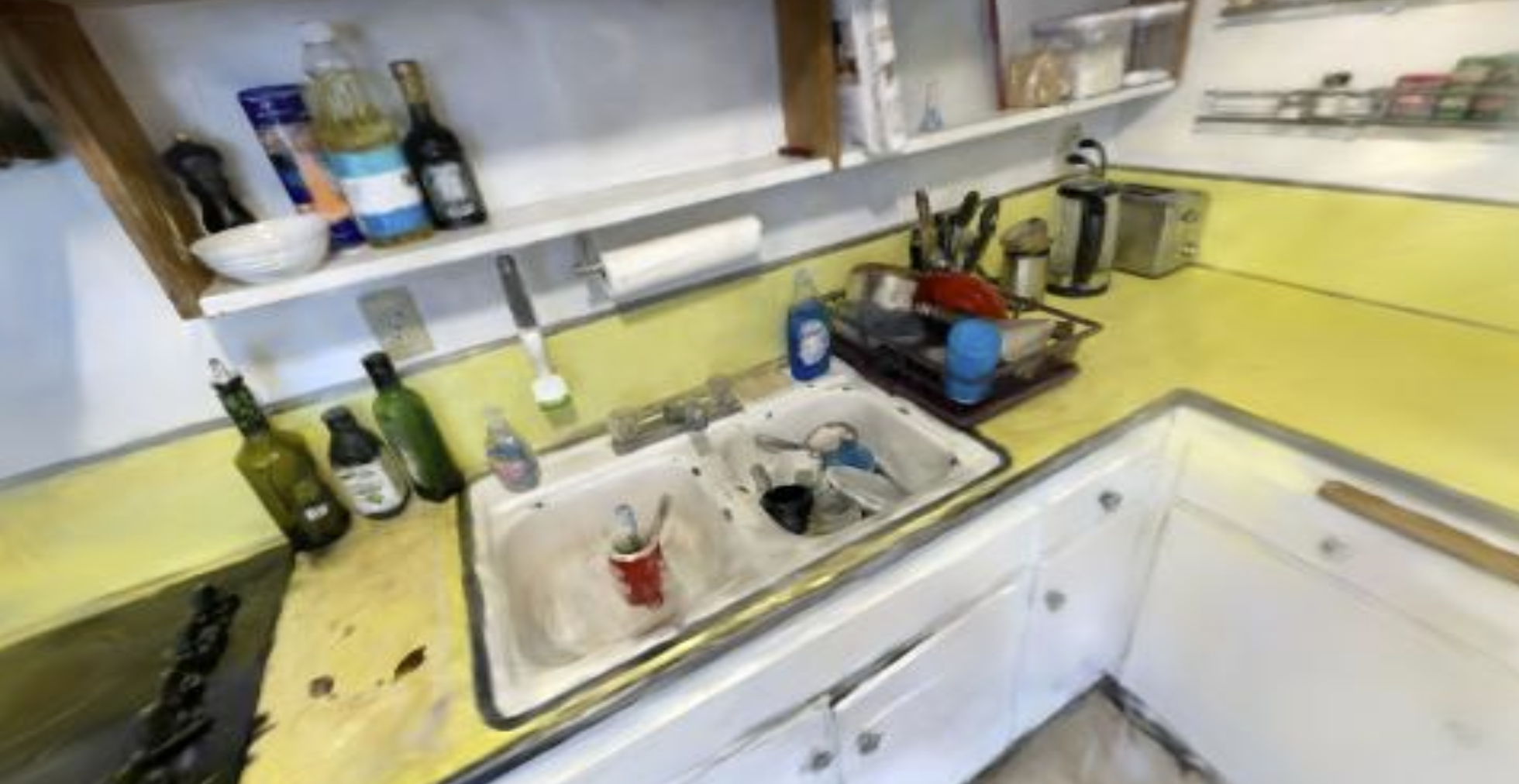}
    \end{minipage}
    \hfill
    \begin{minipage}[t]{0.49\textwidth}
      \centering
      \includegraphics[width=.8\linewidth,height=\hval cm]{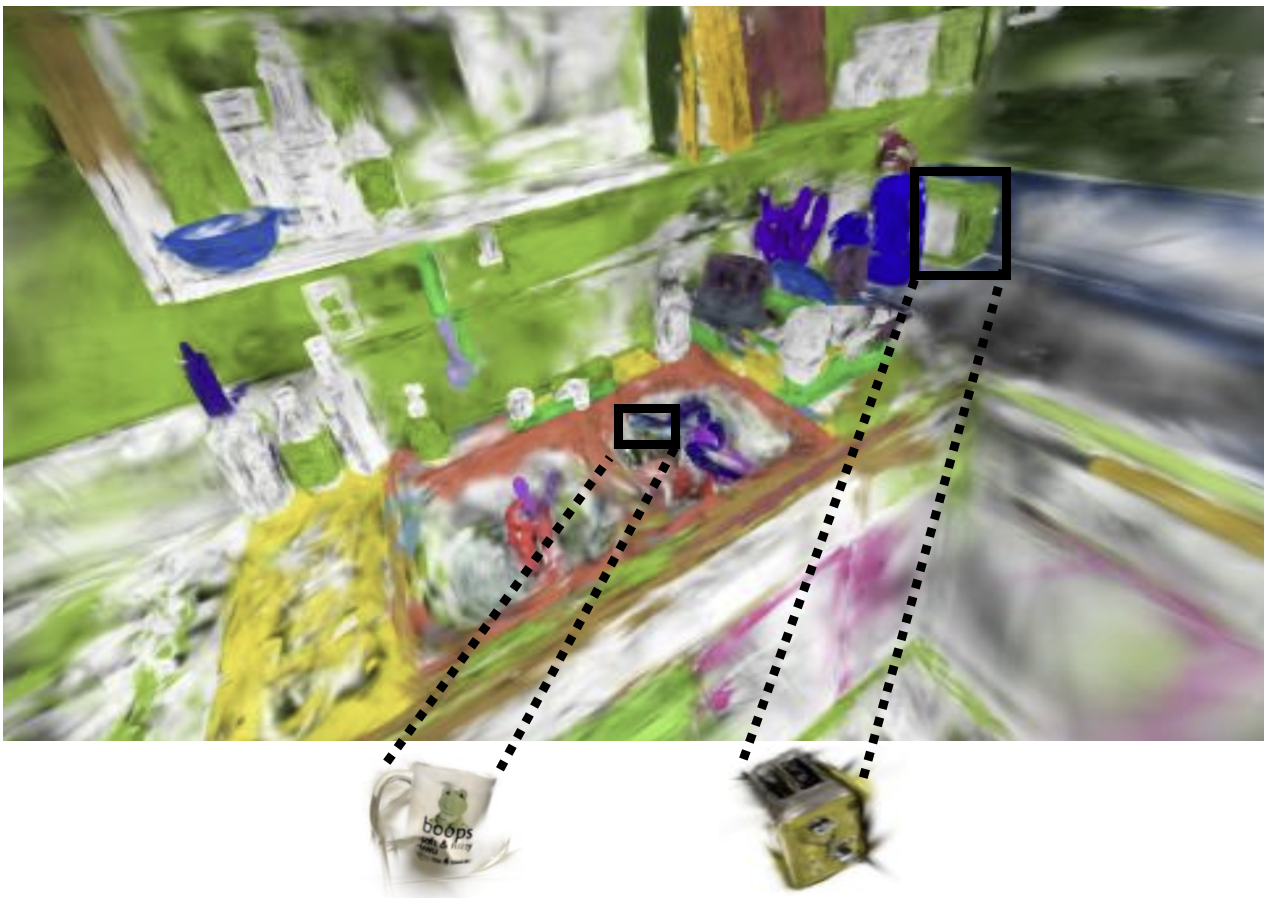}
    \end{minipage}
    \caption{LeRF Kitchen dataset results of final task-relevant objects. Left: RGB rendering, Right: semantic rendering with 
    select extracted 3D objects}
    \label{fig:kitchen_visuals}
\end{figure*}

\newcommand{\hvalp}{4}

\section{Visualizations of Task-Relevant Probabilities}
\label{sec:qualitative_probabilities}

\begin{figure*}[h]
    \centering
    
    \begin{minipage}[t]{0.49\textwidth}
        \centering
        \begin{subfigure}[t]{\textwidth}
            \includegraphics[width=\linewidth,height=\hvalp cm]{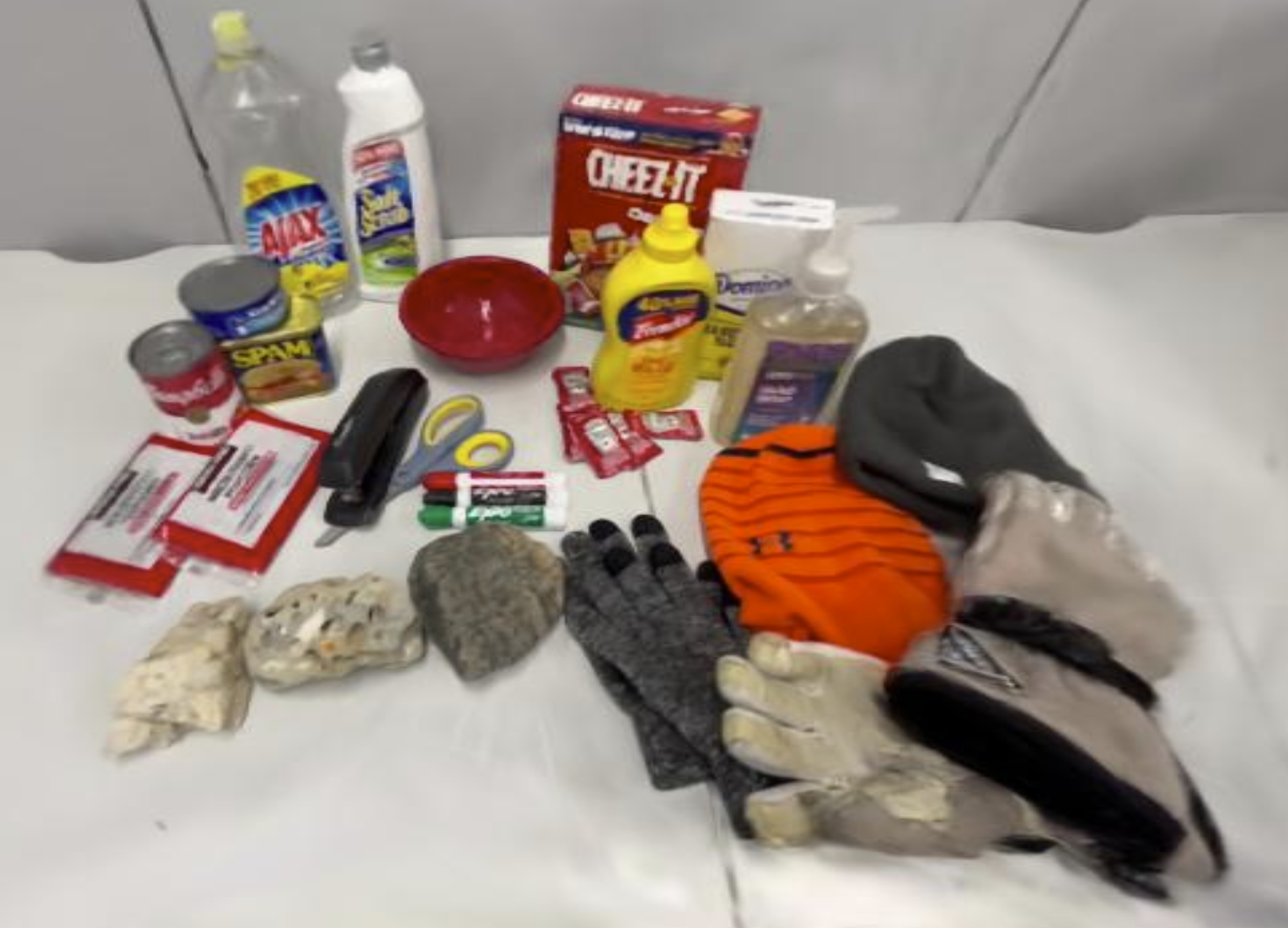}
            \caption{RGB rendering at the same viewpoint of the below images.}
        \end{subfigure}
      \end{minipage}
      \hfill
    
    \vspace{1em}
    \begin{minipage}[t]{0.49\textwidth}
      \includegraphics[width=\linewidth,height=\hvalp cm]{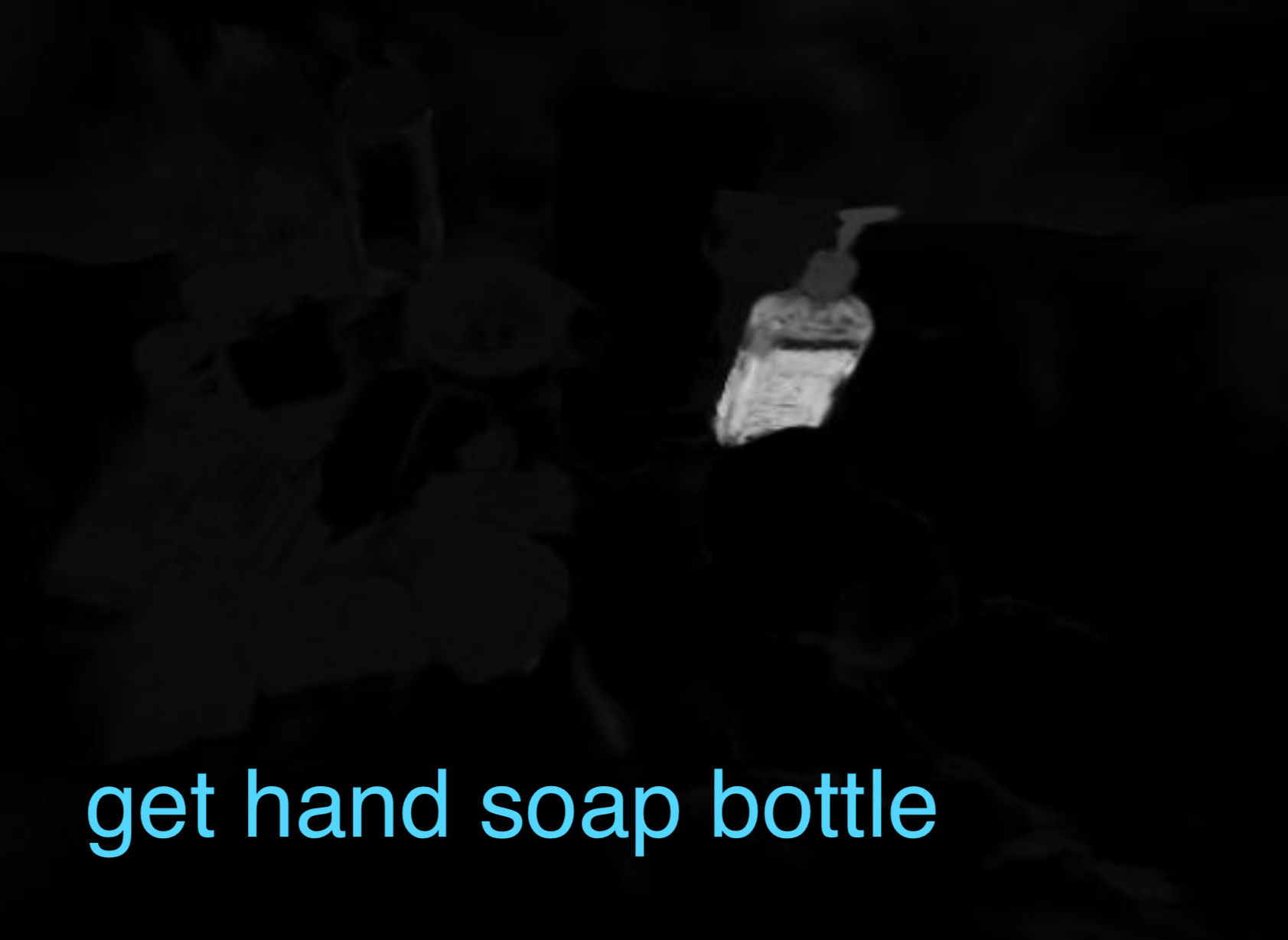}
    \end{minipage}
    \hfill
    \begin{minipage}[t]{0.49\textwidth}
      \includegraphics[width=\linewidth,height=\hvalp cm]{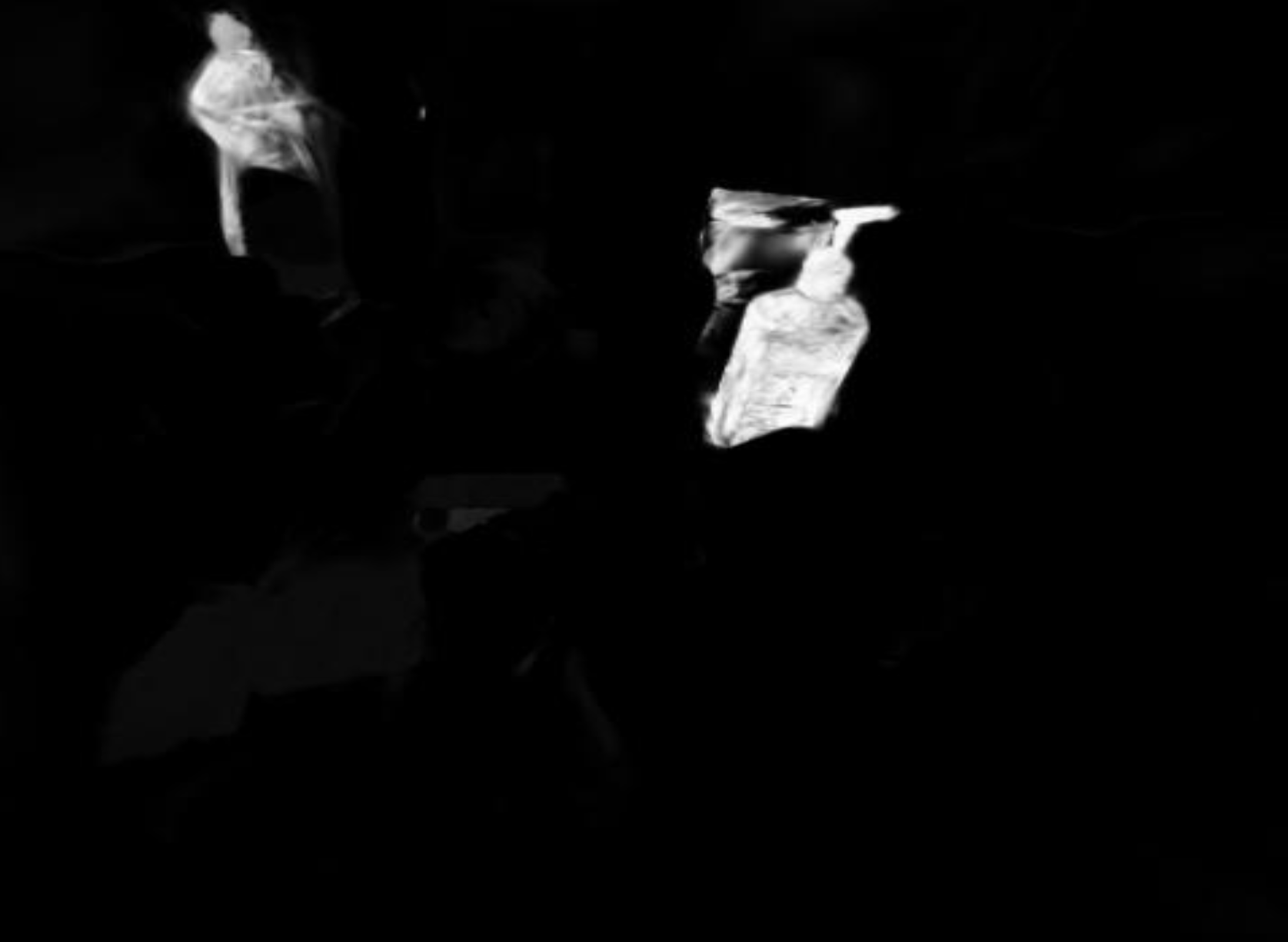}
    \end{minipage}
  
    \vspace{1em}
    \begin{minipage}[t]{0.49\textwidth}
      \includegraphics[width=\linewidth,height=\hvalp cm]{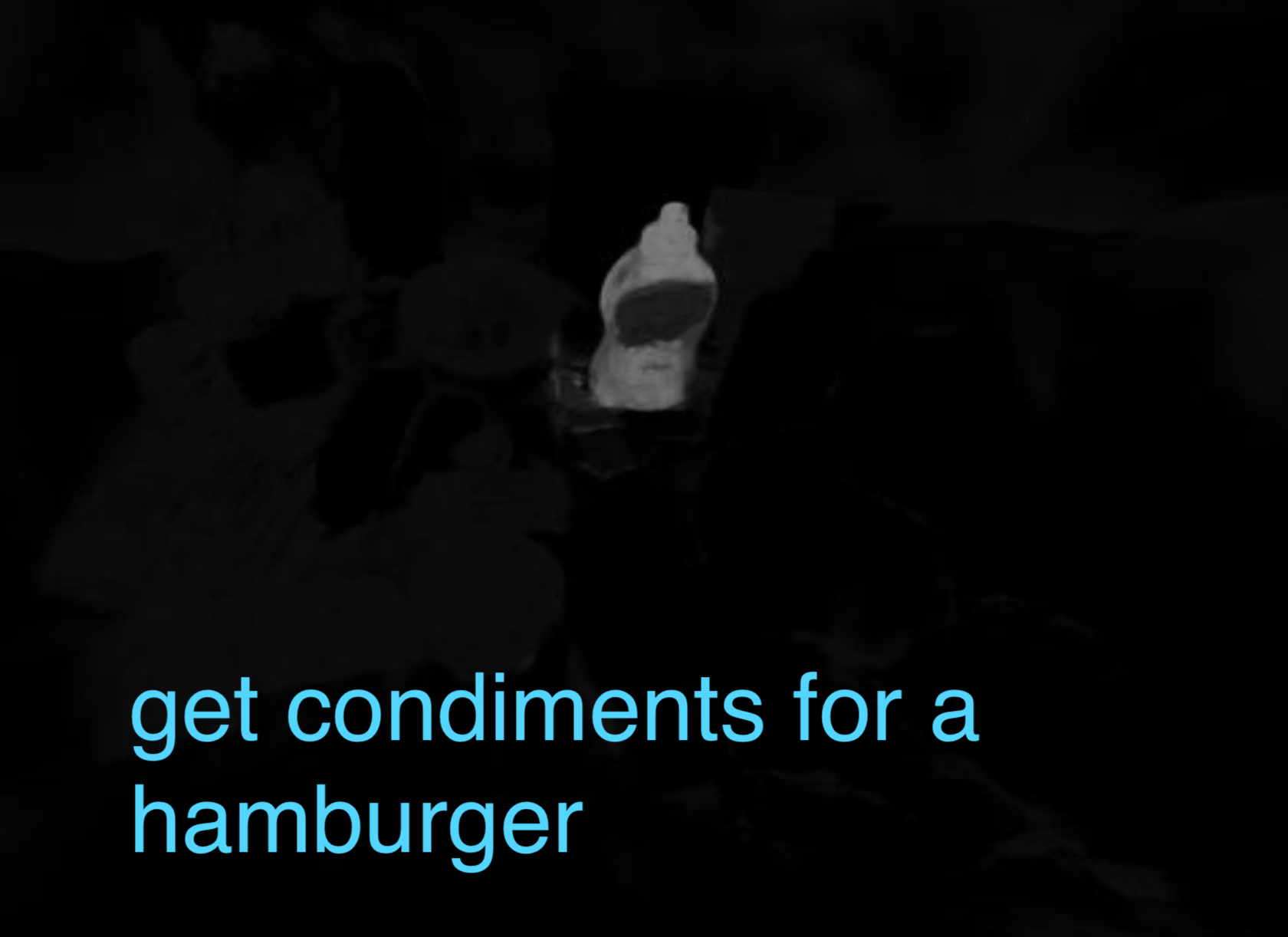}
    \end{minipage}
    \hfill
    \begin{minipage}[t]{0.49\textwidth}
      \includegraphics[width=\linewidth,height=\hvalp cm]{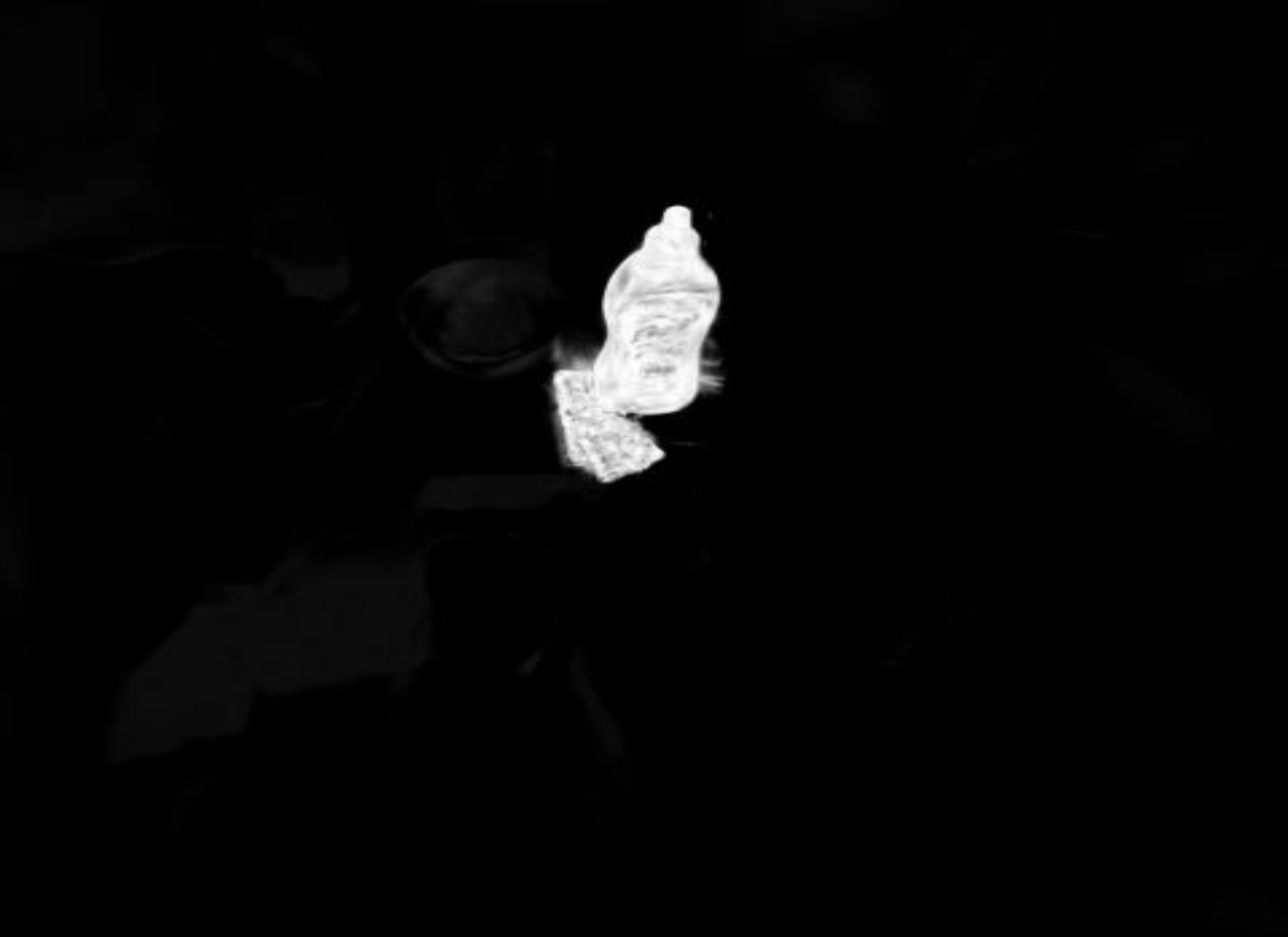}
    \end{minipage}

    \vspace{1em}
    \begin{minipage}[t]{0.49\textwidth}
      \includegraphics[width=\linewidth,height=\hvalp cm]{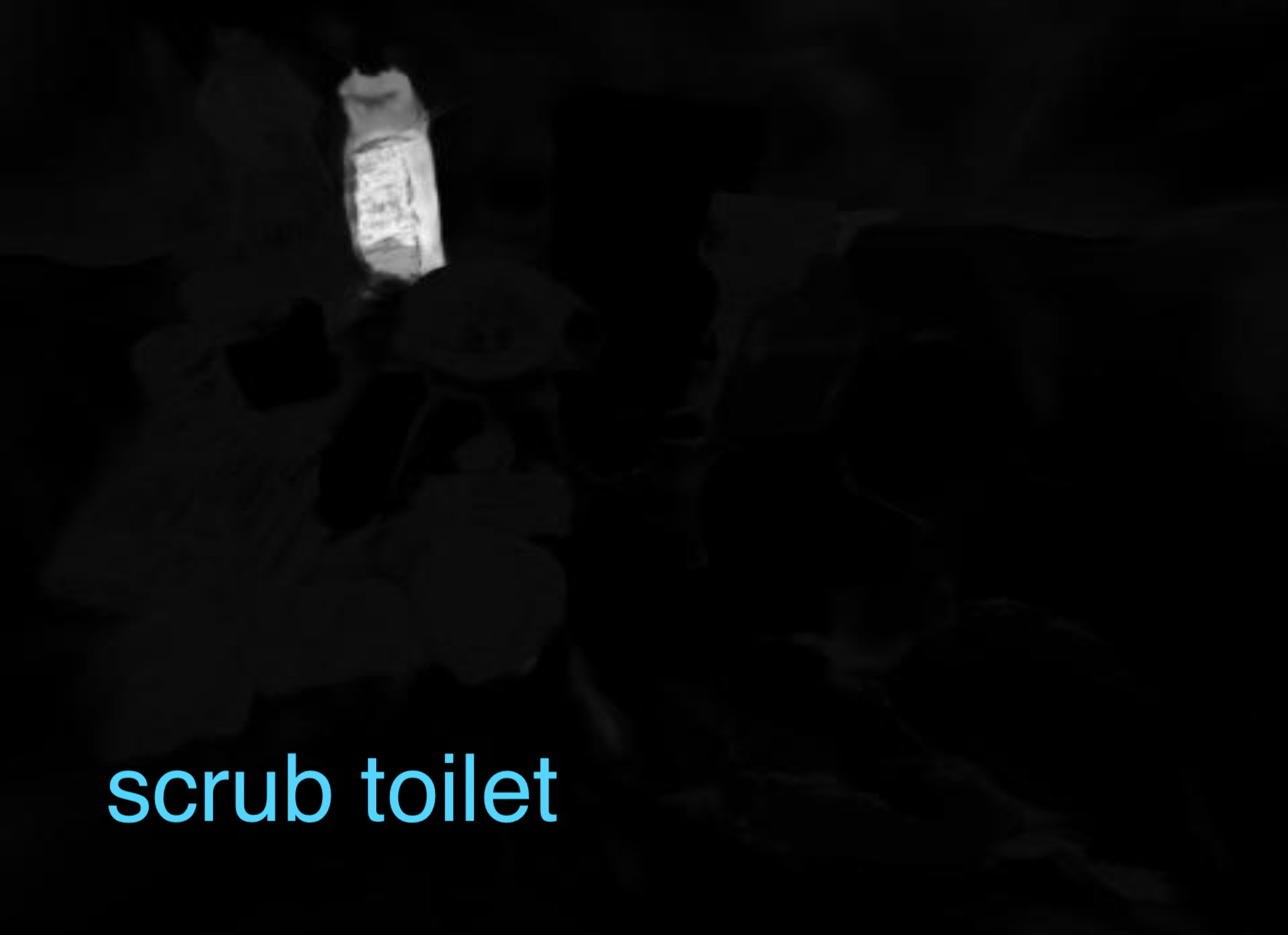}
    \end{minipage}
    \hfill
    \begin{minipage}[t]{0.49\textwidth}
      \includegraphics[width=\linewidth,height=\hvalp cm]{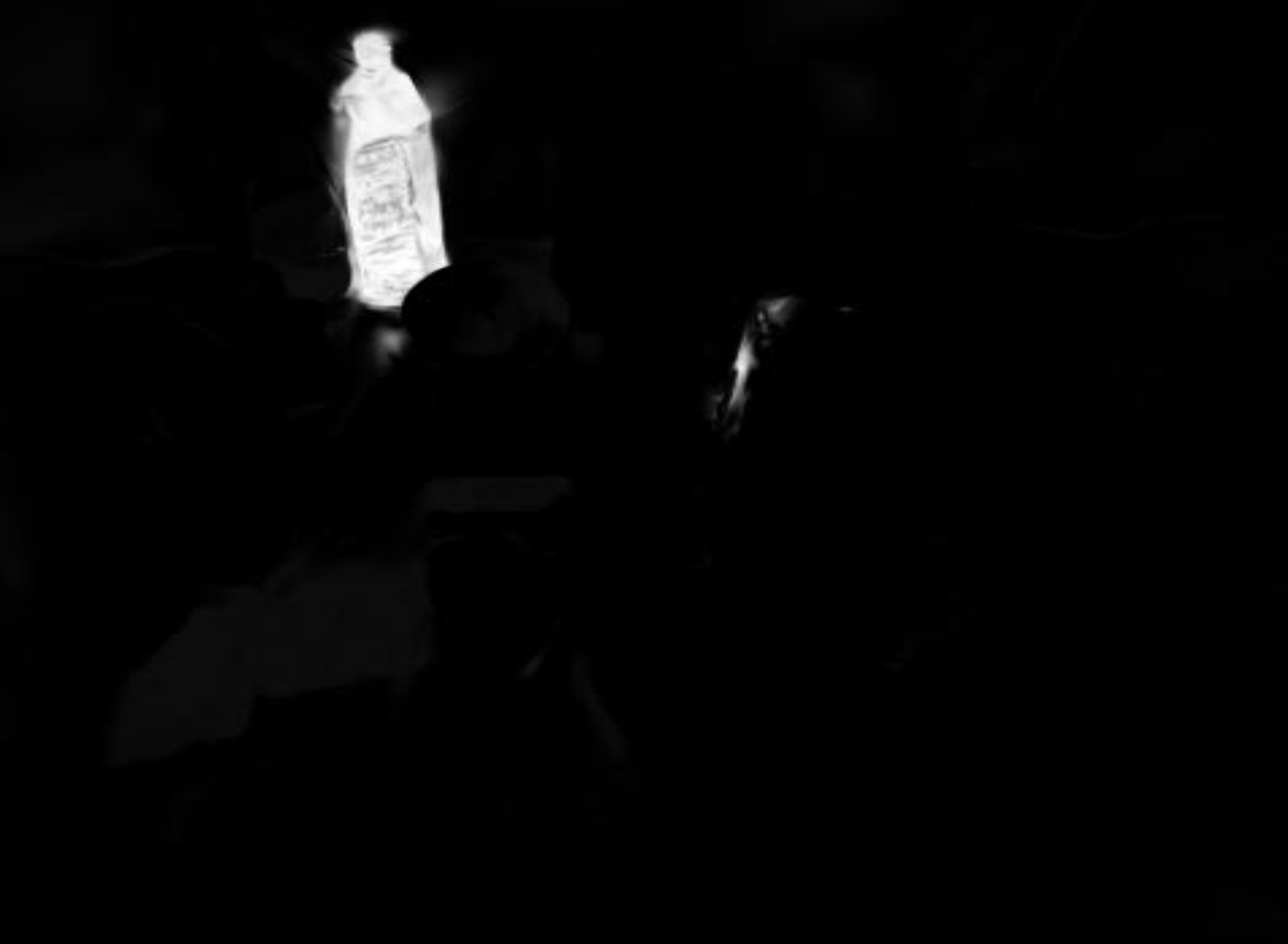}
    \end{minipage}

    \vspace{1em}
    \begin{minipage}[t]{0.49\textwidth}
      \includegraphics[width=\linewidth,height=\hvalp cm]{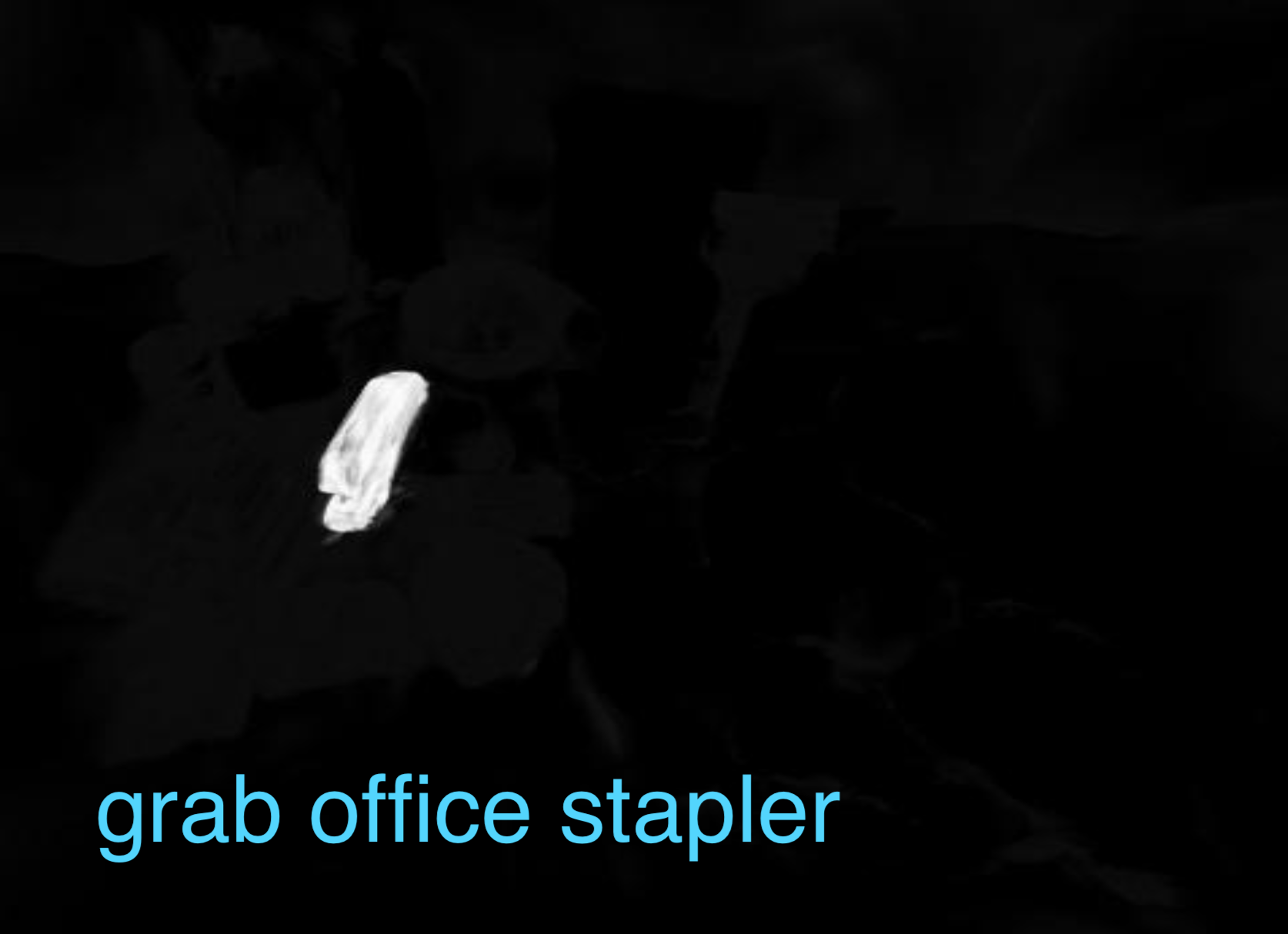}
    \end{minipage}
    \hfill
    \begin{minipage}[t]{0.49\textwidth}
      \includegraphics[width=\linewidth,height=\hvalp cm]{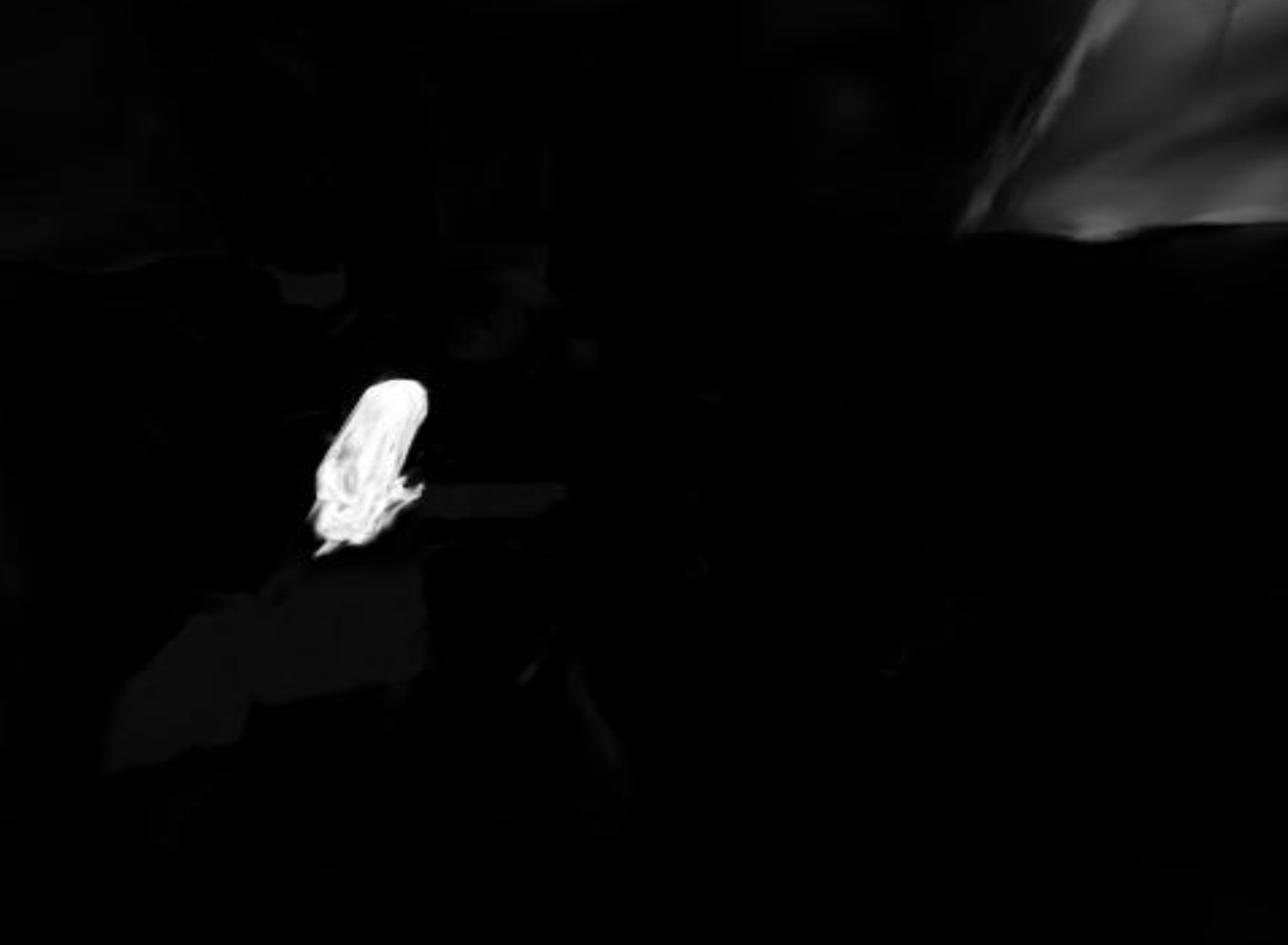}
    \end{minipage}
  
    \caption{Visualization of probabilities to different tasks on 3D Gaussians 
    after a low (left column) and high (right column) number of Bayesian updates. All images 
    are from the same viewpoint.}
    \label{fig:probvisual}
\end{figure*}

To demonstrate the Bayesian updating of probabilities that each 3D Gaussian is relevant for a particular tasks, we include 
a visualization (\cref{fig:probvisual}) of the probability 
assigned to the 3D Gaussians with respect to multiple example tasks for both a few Bayesian updates 
and for all Bayesian updates. For this, we use the Tabletop dataset from \cref{fig:cover_fig}. 
For all examples, the probability of task-relevance to the correct task increases 
after more updates. Additionally, we observe that \name performs well in assigning the relevant Gaussians a high 
probability for each task and the irrelevant Gaussians a low probability. Note this is the probabilities assigned to each Gaussian 
and not the final probabilities assigned to the 3D primitives or the objects. 

\section{Clio Office Dataset}
\label{sec:office_dataset}

As mentioned in \cref{sec:clio_experiments}, we do not include results from running the Clio Office scene as an experiment 
for baseline comparison as the Gaussian Splat 
reconstruction of the Office scene is low quality. However, we include results of running the Office scene in \cref{table:office} 
for completeness. The Office scene contains several doors which have especially poor Gaussian Splat reconstruction while the baseline 
methods using meshes have a better reconstruction of the doors. We include results 
of running on the Office scene with and without tasks involving doors and door handles and notice improvement in the performance of \name 
with these omitted. 
There are still multiple parts of the scene corresponding to tasks that have poor reconstruction, and omitting these further improves performance. 

\begin{table}[h]\scriptsize
    \setlength{\tabcolsep}{2pt}
    \resizebox{\columnwidth}{!}{
    \begin{tabular}{cl ccccc}
    \toprule
    Scene & Method & Strict-osR$\uparrow$ & Relaxed-osR$\uparrow$ & IoU$\uparrow$ & Objs$\downarrow$\\
    \midrule
    & CG~\cite{Gu24icra-conceptgraphs} & 0.24 & 0.52 & 0.07  & 751 \\
    & Khronos~\cite{Schmid24rss-khronos} & \underline{0.67} & 0.67 & \underline{0.15} & 1202 \\
    & Clio-Prim & \textbf{0.70} & \underline{0.73} & \textbf{0.17} & 1883 \\
    \rowcolor{blue!20}\cellcolor{white}& CG-\thres & 0.19 & 0.45 & 0.06  & 40 \\
    \rowcolor{blue!20} \cellcolor{white}& Khronos-\thres & 0.55 & 0.55 & 0.12 & 163 \\
    \rowcolor{blue!20} \cellcolor{white}& Clio-\batch & 0.64 & \textbf{0.76} & 0.13 & 84\\
    \rowcolor{blue!20} \cellcolor{white}& Clio-online & 0.55 & 0.61 & 0.12 & 49 \\
    \rowcolor{blue!20} \cellcolor{white}& \name & 0.55 & 0.67 & \textbf{0.17} & 46 \\
    \midrule
    \multirow{-11}{*}{\rotatebox{90}{Office}}
    & CG~\cite{Gu24icra-conceptgraphs} & 0.26 & 0.52 & 0.07  & 751 \\
    & Khronos~\cite{Schmid24rss-khronos} & \underline{0.70} & 0.70 & \underline{0.17} & 1202 \\
    & Clio-Prim & \textbf{0.74} & \underline{0.78} & \textbf{0.19} & 1880 \\
    \rowcolor{blue!20}\cellcolor{white}& CG-\thres & 0.19 & 0.41 & 0.05  & 35 \\
    \rowcolor{blue!20} \cellcolor{white}& Khronos-\thres & 0.56 & 0.56 & 0.13 & 140 \\
    \rowcolor{blue!20} \cellcolor{white}& Clio-\batch & 0.67 & \textbf{0.82} & 0.13 & 84\\
    \rowcolor{blue!20} \cellcolor{white}& Clio-online & 0.56 & 0.59 & 0.11 & 131 \\
    \rowcolor{blue!20} \cellcolor{white}& \name & 0.67 & 0.70 & \textbf{0.21} & 41 \\
    \midrule
    \multirow{-11}{*}{\rotatebox{90}{Office without doors}}
    \end{tabular}
    }
    \vspace{-4mm}
    \caption{
      Results on the Office scene from the \clio dataset~\cite{Maggio24ral-clio} with and without tasks involving 
      doors and door handles. 
    Highlighted methods have some awareness of the tasks during mapping. 
    \textbf{best}, \underline{second best}
    } 
    \label{table:office}
\end{table}

\clearpage 

\section{Qualitative Comparison with \sg}
\label{sec:sem_gs}
\begin{figure}[h]
  \centering
  \includegraphics[width=0.48\textwidth]{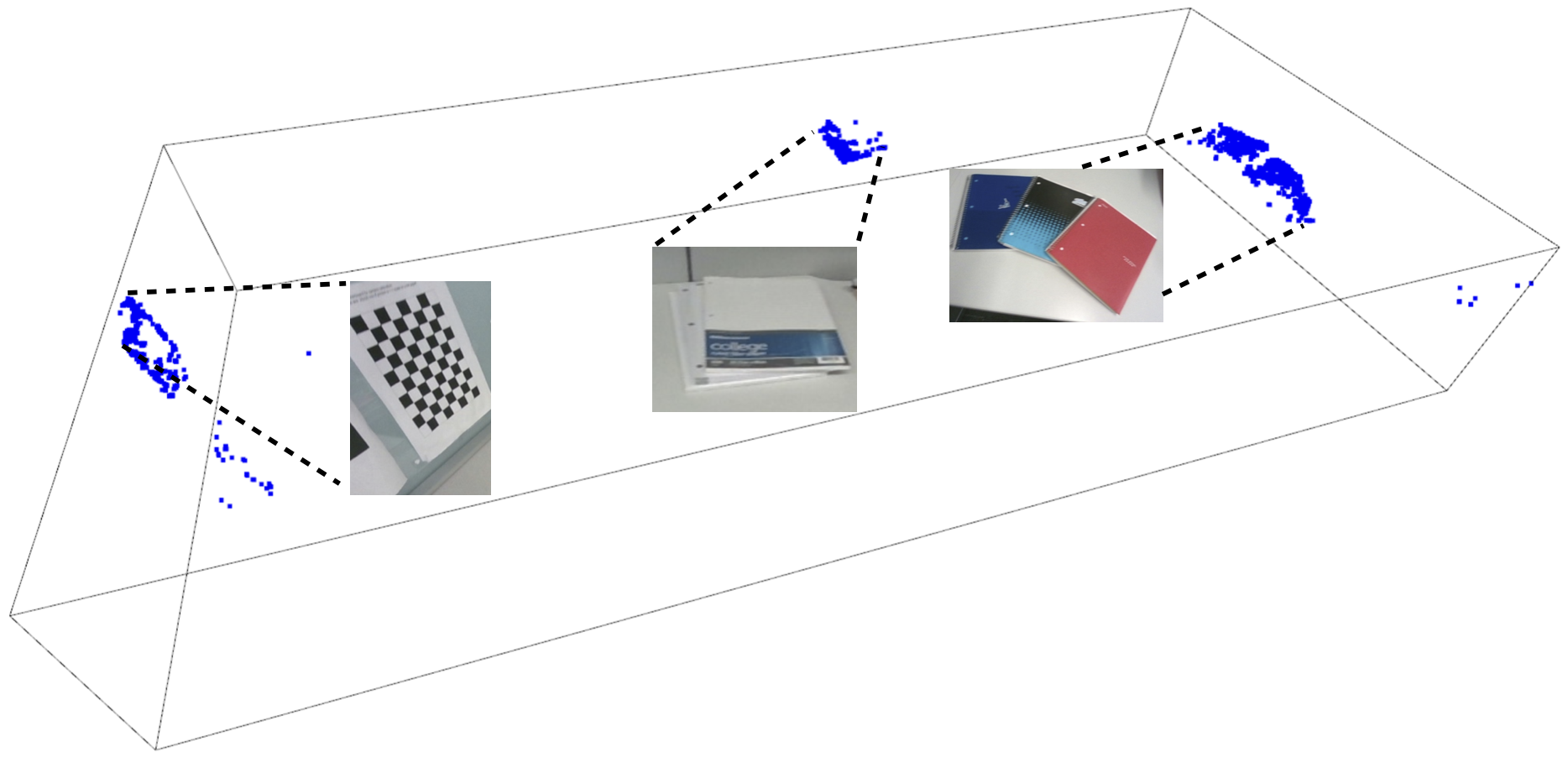}
  \caption{%
      Example of an estimated object for the task ``get notebooks'' from \sg~\cite{Guo24arxiv-semanticGS} on the Clio Cubicle dataset which yields an excessively large object. 
      The visualization includes Gaussians centroids (shown as points) with a minimum 3D oriented bounding box that contains all the points. 
      Images of the scene are shown next to corresponding points for added clarity. This demonstrates that 
      while some Gaussians correspond to the correct 
      region of the scene (the notebooks), other incorrect Gaussians are also returned such as ones from a stack of loose-leaf papers 
      and from paper on the far side of the Cubicle scene.
  }\label{fig:sem_gauss}
\end{figure}

Here we provide a visual explanation of why \sg~\cite{Guo24arxiv-semanticGS} has a low IoU score for the Clio dataset experiments (\cref{table:custom_datasets}). 
For example, in \cref{fig:sem_gauss} for the task ``get notebooks'', 
\sg correctly returns Gaussians associated with the notebooks, but also returns Gaussians 
associated with other regions of the scene such as 
from a stack of papers. This results in a low IoU score for \sg in \cref{sec:clio_experiments} as the resulting 3D bounding box for 
many of the estimated objects such as 
this are overly large.

\section{Qualitative Comparison with Garfield}
\label{sec:garfield}

\begin{figure*}[h]
  \centering
  \begin{subfigure}[t]{0.48\linewidth}
    \centering
    \includegraphics[width=\linewidth]{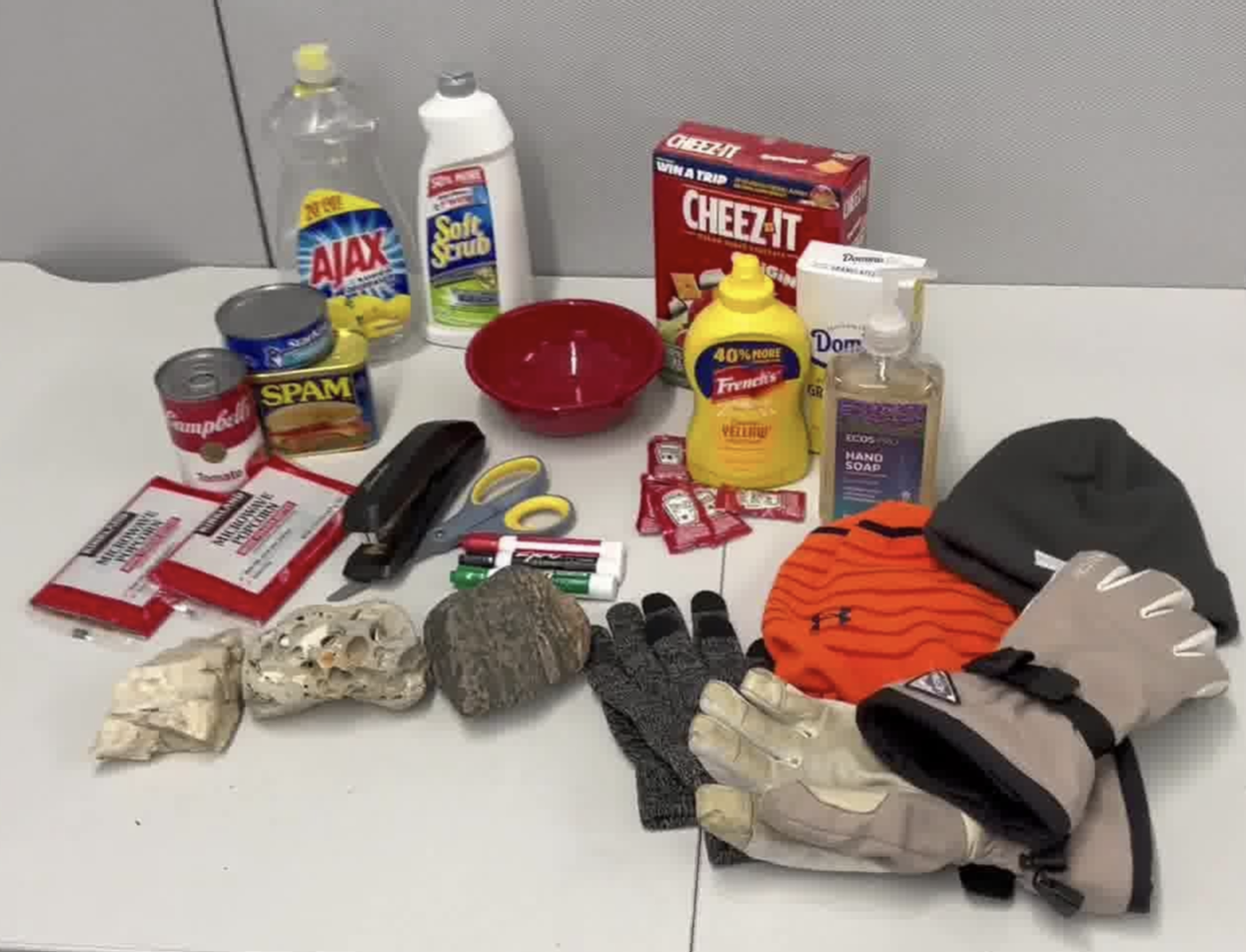}
    \caption{RGB view of Tabletop experiment scene used in \cref{fig:cover_fig}.}
    \label{fig:cam_setup}
  \end{subfigure}
  \hfill
  \begin{subfigure}[t]{0.48\linewidth}
    \centering
    \includegraphics[width=\linewidth]{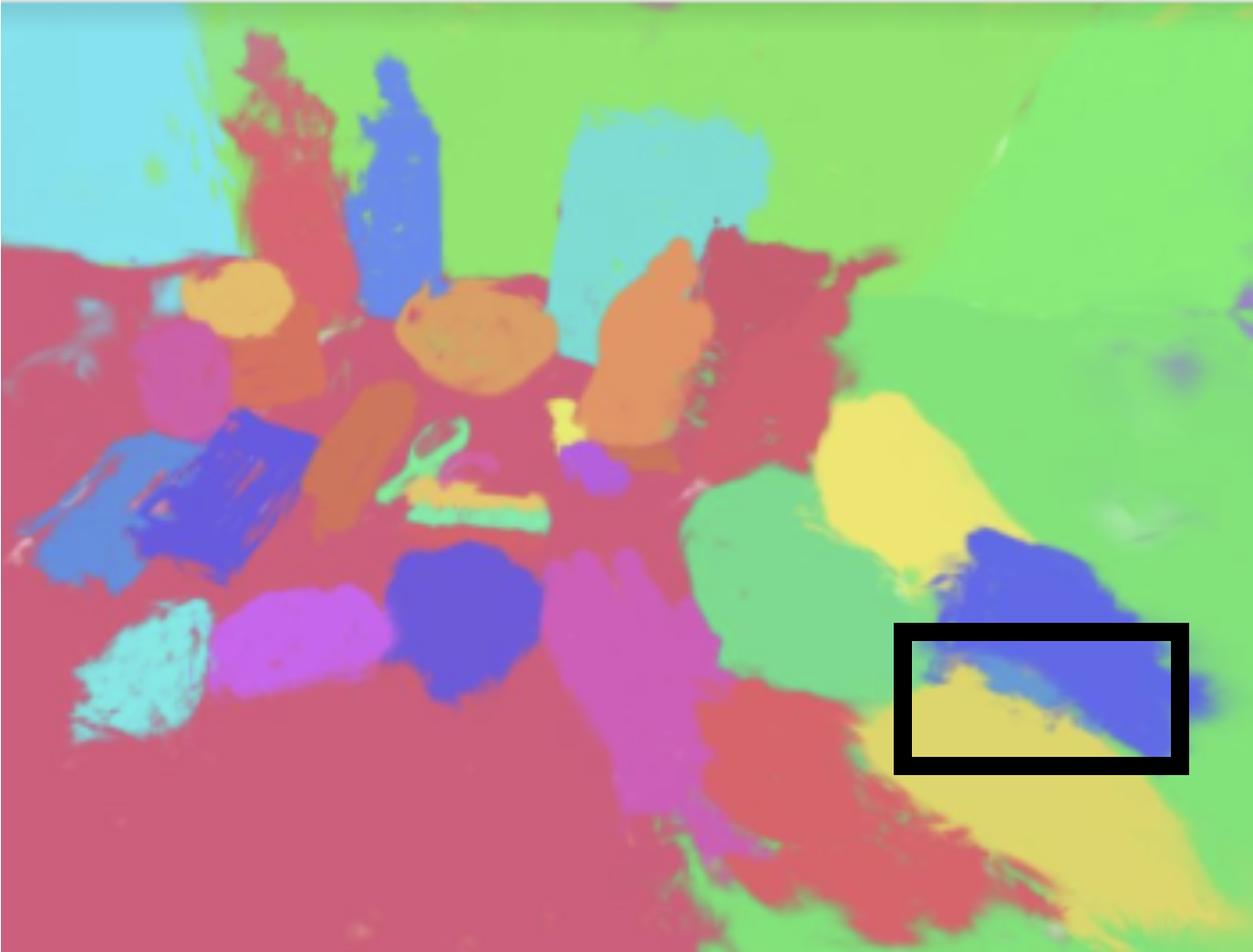}
    \caption{Garfield clustering with fine granularity. The buckle is labeled with a bounding box.}
    \label{fig:garfield_fine}
  \end{subfigure}
  
  \vskip\baselineskip
  
  \begin{subfigure}[t]{0.48\linewidth}
    \centering
    \includegraphics[width=\linewidth]{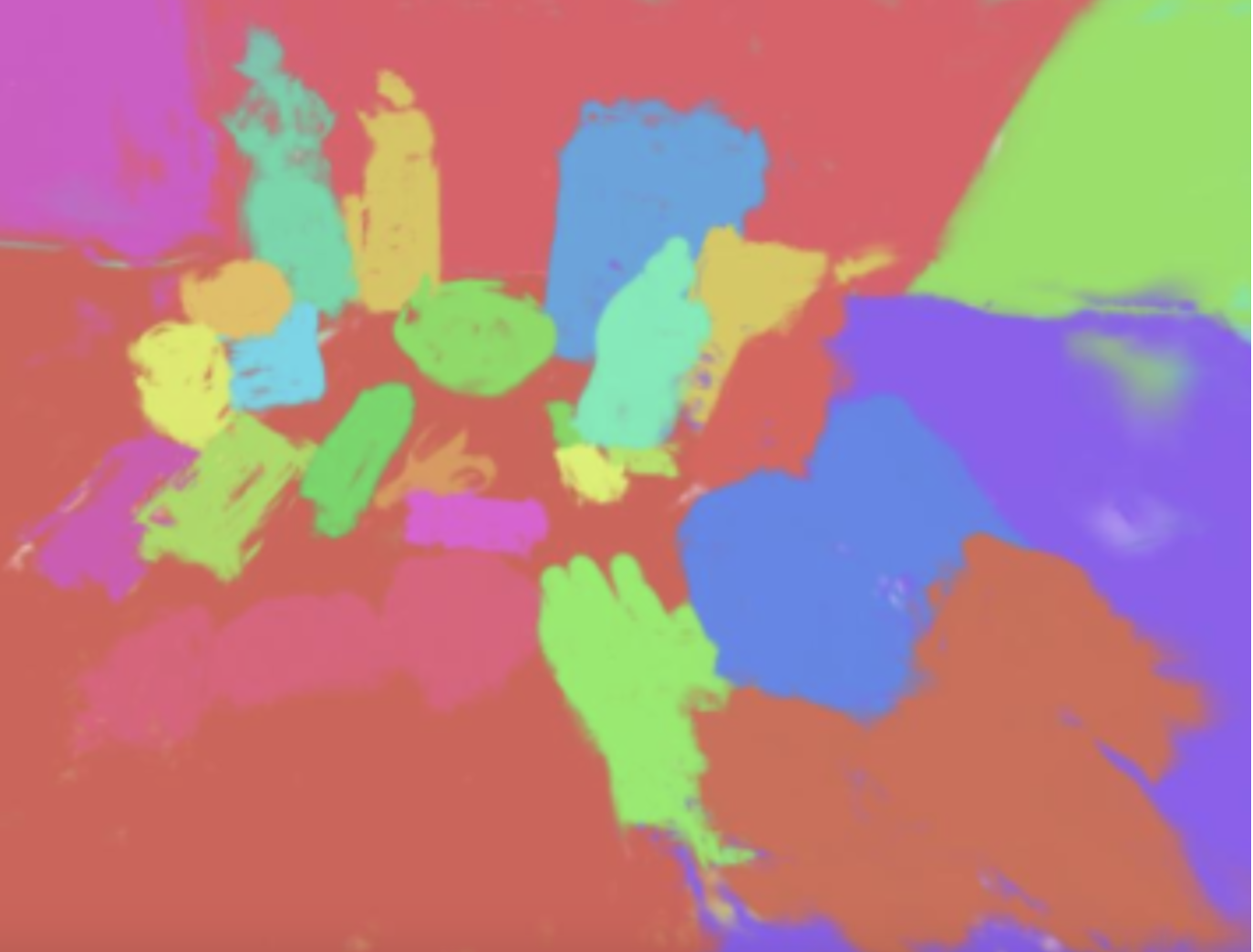}
    \caption{Garfield clustering with medium granularity.}
    \label{fig:garfield_medium}
  \end{subfigure}
  \hfill
  \begin{subfigure}[t]{0.48\linewidth}
    \centering
    \includegraphics[width=\linewidth]{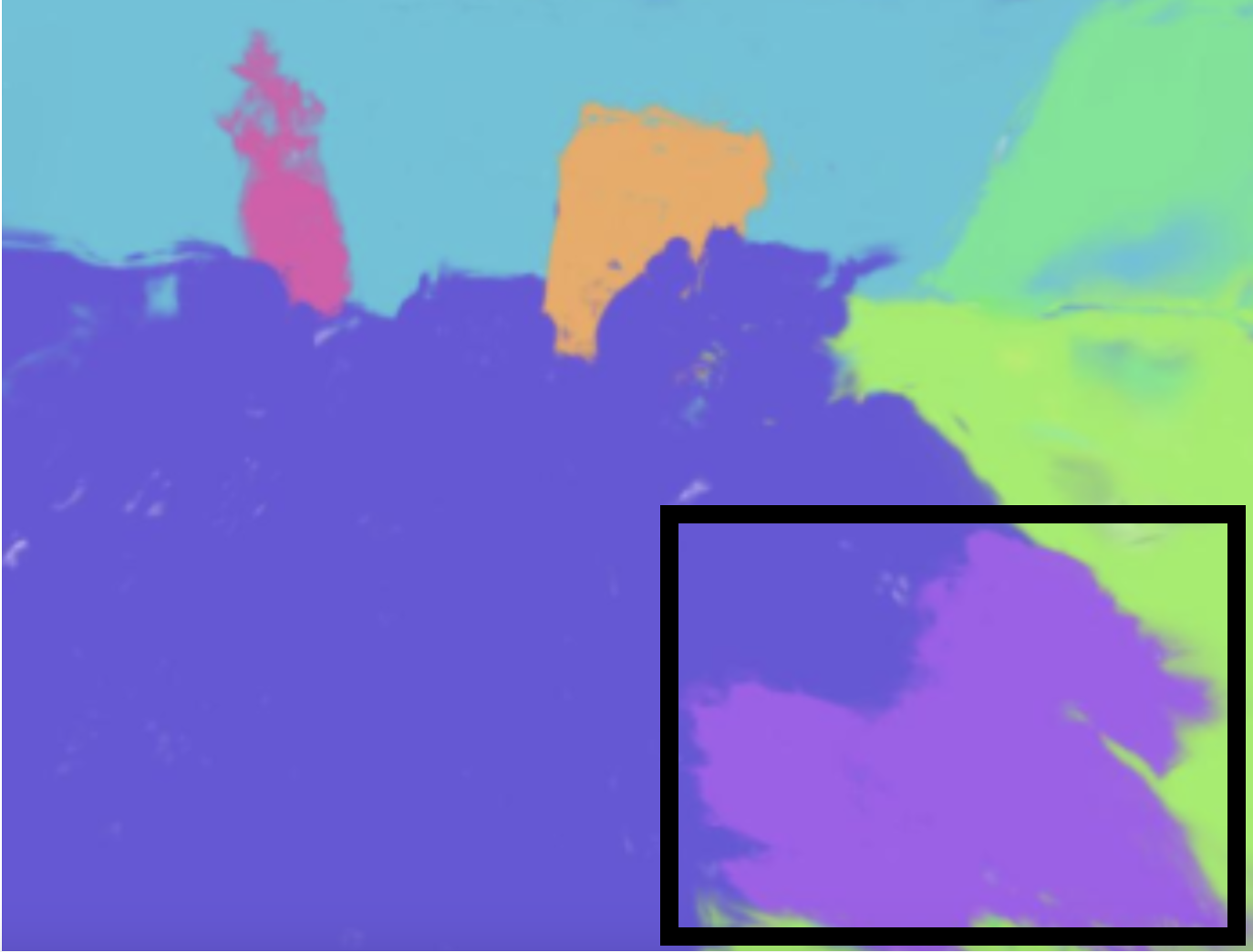}
    \caption{Garfield clustering with coarse granularity. The pile of clothes is labeled with a bounding box.}
    \label{fig:garfield_course}
  \end{subfigure}
  
  \caption{Qualitative clustering results of running Garfield~\cite{Kim24cvpr-garfield} on our Tabletop scene.}
  \label{fig:garfield}
\end{figure*}

As mentioned in \cref{sec:related_work}, Garfield~\cite{Kim24cvpr-garfield} uses a trained affinity field based on masks from 
SAM~\cite{Kirillov23iccv-SegmentAnything} to cluster a scene at differently granularity levels. Since semantics are not included in 
the Garfield map and since granularity is determined with a user set parameter, we generate qualitative clustering 
results from Garfield by clustering 
the entire scene from the Tabletop dataset in \cref{fig:cover_fig} at three different 
granularity levels (\cref{fig:garfield}). 
The finest granularity (\cref{fig:garfield_fine}) is set as the largest granularity that considers the buckle on the gloves as an object. This 
is the correct object for the task ``get buckle'' (task A4 in \cref{fig:cover_fig}). 
The coarsest granularity (\cref{fig:garfield_course}) 
is set as the smallest granularity which considers as much of the 
pile of clothes as possible as one object. This is the correct object for the task ``move hats and gloves'' (task B2 in \cref{fig:cover_fig}). 
Note however, that no granularity with Garfield correctly 
captured the entire pile. 
We also include a medium granularity (\cref{fig:garfield_medium}). 
For the fine granularity, while the resolution is correct for the buckle, 
other regions of the scene are too fine. Likewise, while some of the pile of clothes is correctly identified in the course granularity, 
the remainder of the scene is too coarse. 
This emphasizes the importance of representing different regions of the scene at different 
granularities. 
As we keep all parameters for \name constant for all experiments, 
we are unable to create valid clustering with Garfield on the 
larger Apartment scene using the parameters required for training the Tabletop scene.

\end{document}